\documentclass[11pt]{article}

\usepackage{microtype} 
\usepackage{booktabs}  

\usepackage{amsmath}
\usepackage{amsthm}
\usepackage{xspace}

\usepackage[final]{automl}

\usepackage[square,numbers,compress]{natbib}
\usepackage{wrapfig}
\usepackage{nicefrac}
\usepackage{algorithm}
\usepackage[noend]{algpseudocode}
\usepackage{color}
\usepackage{tikz}
\usetikzlibrary{positioning}

\newcommand{\sectionSpace}{\vspace{-0.0075\textheight}}
\newcommand{\sectionSpaceAfter}{\vspace{-0.005\textheight}}

\newcommand{\subsectionSpace}{\sectionSpace}

\newcommand{\dev}{dev.\xspace}
\renewcommand{\prod}{prod.\xspace}
\newcommand{\Algorithm}{Alg.\xspace}
\newcommand{\Figure}{Fig.\xspace}

\title{Task Selection for AutoML System Evaluation}

                    \author[1]{\nameemail{\small{Jonathan Lorraine}}{jonlorraine@google.com}}
                    \author[1]{\nameemail{\small{Nihesh Anderson}}{nihesh@google.com}}
                    \author[1]{\nameemail{\small{Chansoo Lee}}{chansoo@example.com}}
                    \author[1]{\nameemail{\small{Quentin De Laroussilhe}}{underflow@google.com}}
                    \author[1]{\nameemail{\small{Mehadi Hassen}}{mehadi@example.com}}

                    \affil[1]{Google}

\hypersetup{%
  pdfauthor={}, 
  pdftitle={},
  pdfsubject={},
  pdfkeywords={}
}

\begin{document}
    \maketitle
    \begin{abstract}
        Our goal is to assess if AutoML system changes - i.e., to the search space or hyperparameter optimization - will improve the final model’s performance on production tasks.
        However, we cannot test the changes on production tasks.
        Instead, we only have access to limited descriptors about tasks that our AutoML system previously executed, like the number of data points or features.
        We also have a set of development tasks to test changes, ex., sampled from OpenML with no usage constraints.
        However, the development and production task distributions are different leading us to pursue changes that only improve development and not production. 
        This paper proposes a method to leverage descriptor information about AutoML production tasks to select a filtered subset of the most relevant development tasks.
        Empirical studies show that our filtering strategy improves the ability to assess AutoML system changes on holdout tasks with different distributions than development.
    \end{abstract}

    \vspace{-0.03\textheight}
    \section{Introduction} \label{sec:intro}
    \sectionSpaceAfter
    \begin{wrapfigure}[15]{r}{0.54\textwidth}
            \centering
            \vspace{-0.05\textheight}
            \begin{tikzpicture}
                \centering
                \node (img){\includegraphics[trim={.75cm .25cm .65cm 1.15cm},clip, width=.9\linewidth]{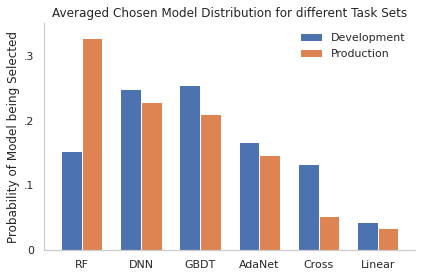}};
                \node[left=of img, node distance=0cm, rotate=90, xshift=1.3cm, yshift=-.85cm, font=\color{black}] {\small{Prob.~of selection}};
                \node[below=of img, node distance=0cm, xshift=.2cm, yshift=1.1cm,font=\color{black}] {\small{Selected model type}};
            \end{tikzpicture}
            \vspace{-0.0175\textheight}
            \caption{
                We show the distribution of selected model types for the \prod tasks and \dev tasks.
                The choices are random forest (RF), deep neural networks (DNN), gradient-boosted decision tree (GBDT), AdaNet, Linear-Feature-Cross (Cross), and a linear model.
            }\label{fig:prod_model_selection}
            \vspace{-0.02\textheight}
        \end{wrapfigure}
        Successful deployment of machine learning models requires many design choices, often requiring expertise.
        Automatic Machine Learning (AutoML) systems mechanize this process.
        A typical AutoML system consists of many components, including feature engineering, model selection \& optimization~\cite{zoller2021benchmark, arango2021hpo, eggensperger2021hpobench}, and hyperparameter optimization~\cite{bergstra2011algorithms, hutter2011sequential, snoek2015scalable, springenberg2016bayesian}.
        
        Providers usually deploy an AutoML system as a service, which they want to improve by rolling out changes and assessing system performance.
        Due to providers limited visibility into \emph{production tasks} (abbreviated \prod) supplied by actual clients, it is impossible to run system experiments on these tasks.
        Instead, providers have a set of open-sourced \emph{development tasks} (abbreviated \dev)~\cite{OpenML2013, dataset_openml_adult, dataset_openml_bank_marketing, dataset_nomao, dataset_madelon, dataset_wilt, dataset_civil_comments, dataset_yelp_sentiment, dataset_criteo} which can be used to assess system changes.
        
\begin{figure}[H]
            \centering
            \vspace{-0.05\textheight}
            \begin{tikzpicture}
                \centering
                \node (img){\includegraphics[trim={.75cm .8cm .6cm 1.0cm},clip, width=.46\linewidth]{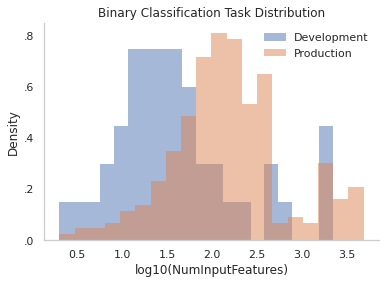}};
                \node[left=of img, node distance=0cm, rotate=90, xshift=1.0cm, yshift=-.9cm, font=\color{black}] {\small{Density}};
                \node[below=of img, node distance=0cm, xshift=-.0cm, yshift=1.25cm,font=\color{black}] {\footnotesize{$\log_{10}$ of number of input features}};
                
                \node [right=of img, xshift=-1.1cm](img2){\includegraphics[trim={1.25cm .8cm .6cm 1.0cm},clip, width=.44\linewidth]{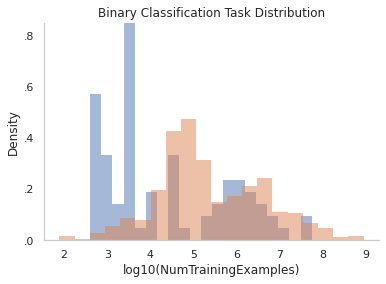}};
                \node[below=of img2, node distance=0cm, xshift=-.0cm, yshift=1.25cm,font=\color{black}] {\footnotesize{$\log_{10}$ of number of data points}};
            \end{tikzpicture}
            \vspace{-0.015\textheight}
            \caption{
                We show histograms of features for both the \prod and \dev task distributions, illustrating significant differences.
            }\label{fig:distribution_shift}
            \vspace{-0.02\textheight}
        \end{figure}
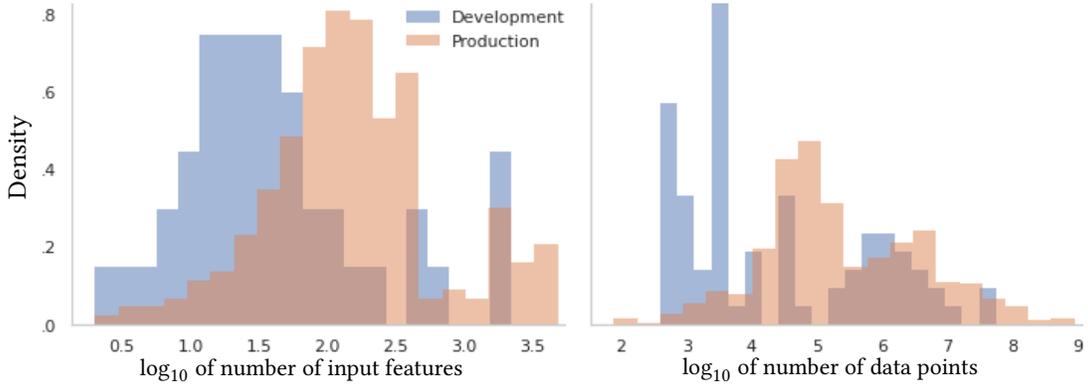

        However, the distribution of the \dev tasks may be wildly different from the received \prod tasks -- see (App.)endix \Figure~\ref{fig:prod_model_selection} and \ref{fig:distribution_shift}.
        \Figure~\ref{fig:prod_model_selection} displays the distribution of selected model types for a set of \prod and \dev tasks, showing a significant difference, which could -- for example -- lead us to pursue DNN improvements when a random forest is used most often in prod.
        Therefore, testing the impact of a change on real users is a non-trivial task unaddressed by existing papers on benchmarking~\cite{zoller2021benchmark, arango2021hpo, eggensperger2021hpobench, bischl2017openml, balaji2018benchmarking, elshawi2019automated}.
        %
        Our contributions in this paper include:
        \begin{enumerate}
            \item \sectionSpaceAfter Proposing a framework to select a filtered subset of \dev tasks so the performance delta -- on a change to our AutoML system -- is similar between filtered and \prod tasks.
            \item \sectionSpace Proposing task selection methods and comparing their trade-offs in large-scale experiments.
            \item \sectionSpace Showing specific examples of methods which improve our ability to accurately assess AutoML system changes when we have task distribution shift, as in production setups.
            \sectionSpace
        \end{enumerate}

    \section{Background and Key Concepts} \label{sec:background}
        Notations used in the paper are summarized in Appendix Table~\ref{tab:TableOfNotation}. 

        \textbf{The Production AutoML System:}\label{sec:our_autoML_system}
            The key components of our AutoML system include the search space, hyperparameter optimizer, and how we assess performance.
            The AutoML system is a service used on varied machine learning tasks for commercial applications, supporting multi-modal datasets with image, natural language text, and tabular data.
            Specific details are in App. Sec.~\ref{sec:app_our_automl_system_results}, summarized here:
            We use a conditional search space, with a root parameter for the algorithm choice (\Figure~\ref{fig:prod_model_selection}) and sub-parameters conditioned on the root choice.
            A wide range of hyperparameters are tuned, including optimizer, model choice, regularization, and feature engineering.
            Vizier~\cite{golovin2017google} is our hyperparameter optimizer, supporting user-specified parameters controlling the HPO.
        
        \textbf{Changes to the AutoML System:}\label{sec:our_changes}
            AutoML service providers constantly make \emph{changes} to the AutoML system in order to improve performance.  
            By \emph{change}, we mean moving from an AutoML system with a \emph{baseline setup} to a \emph{modified setup}.
            Generally, we can view each setup as an arbitrary AutoML system; thus, the change can encode almost any reasonable system modification.
            System changes can be partitioned into those affecting the AutoML system/search space, and those for the hyperparameter optimizer.
            We look at both in our experiments in Sec.~\ref{sec:filter_setup}.
            
        \textbf{Tasks:}\label{sec:tasks}
            AutoML service providers receive a stream of tasks from their end-users.
            Here, a \emph{task} is a dataset and a \emph{problem statement} specifying what is to be solved and the relevant metric.
            For example, stating that we have a binary classification problem for some specified label, and the goal is maximizing AUC.
            Our systems run on two sets of tasks:
            \emph{Development tasks} (dev.) are gathered from various open-source repositories (such as OpenML \cite{OpenML2013}), have no restrictions and thus can be easily used to evaluate our service.
            The second set is \emph{Production tasks} (prod.), which are those received by our service from a client and \emph{can only be run for the exact client-specified purpose}.
            This setup has parallels with meta-learning (ex., task splits), but with additional issues and information/constraints available for providers.

        \textbf{Unique Production Issues for AutoML Service Providers:}\label{sec:service_provider_issue}
            AutoML service providers have limited visibility into the uses of the ML system in product applications.
            This could, for example, be to comply with privacy regulations and the terms of service.
            Similarly, providers may not want to risk evaluating experiments on model changes in live production environments.
            These constraints force providers to assess system performance on \dev tasks.

        \textbf{System descriptors:}
            AutoML service providers often have access to common \emph{descriptors} about the executed tasks.
            These include \emph{task descriptors}, containing info about only the task, such as the number of data points or number/modality of features.
            Alternatively, \emph{system descriptors} contain info about both the task and our AutoML system.
            For example, the performance (ex., accuracy) of a model on a task with some hyperparameter values.
     
    \newpage
    \begin{figure}[H]
            \centering
            \vspace{-0.05\textheight}
            \begin{tikzpicture}
                \centering
                \node (img){\includegraphics[trim={.0cm .0cm .0cm .0cm},clip, width=.47\linewidth]{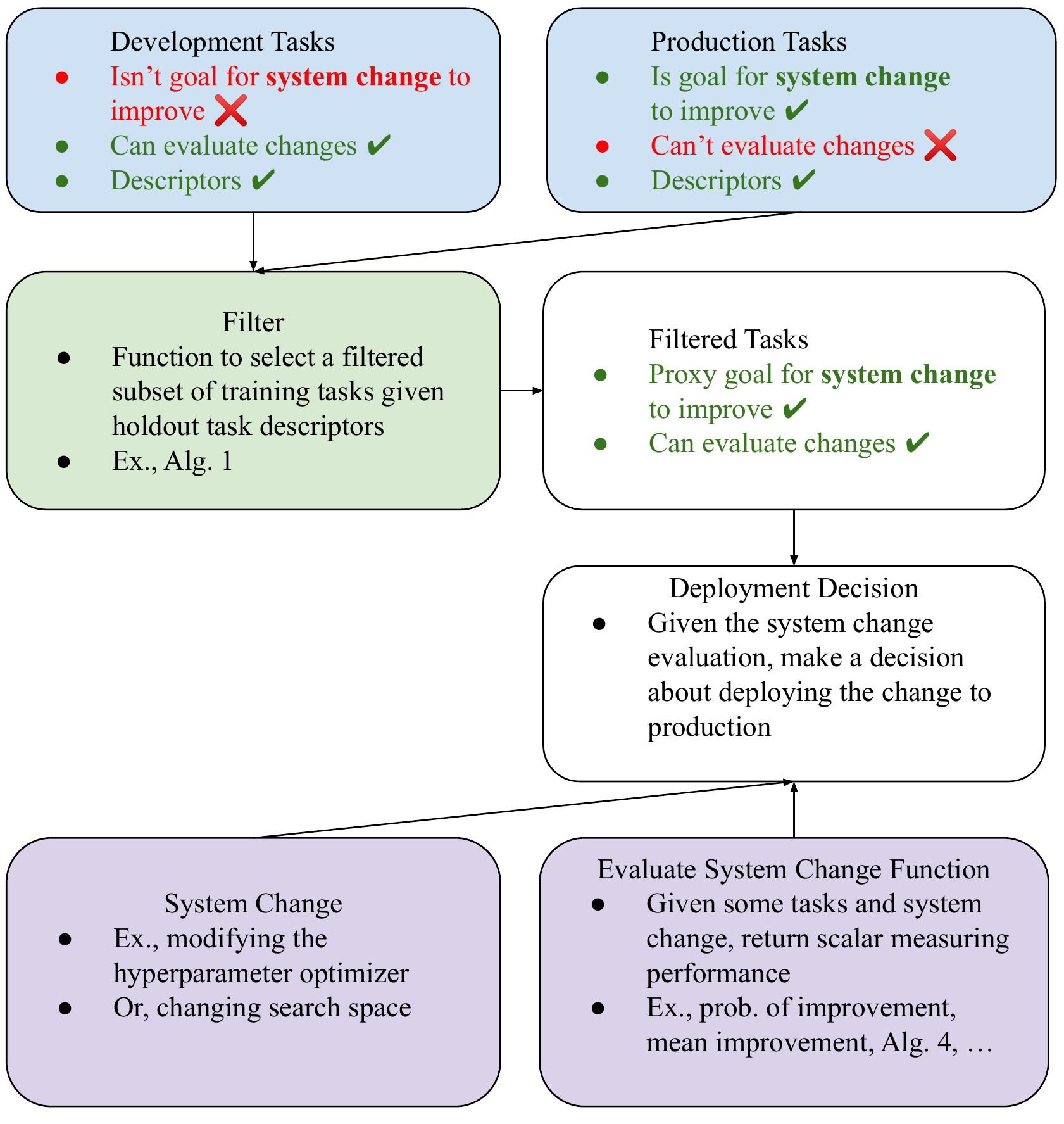}};
                \node[above=of img, node distance=0cm, xshift=-.0cm, yshift=-1.2cm,font=\color{black}] {\footnotesize{{\color{red}Using} Filters for Production Deployment Decisions}};
                
                \node [right=of img, xshift=-1cm](img2){\includegraphics[trim={.0cm .0cm .0cm .0cm},clip, width=.47\linewidth]{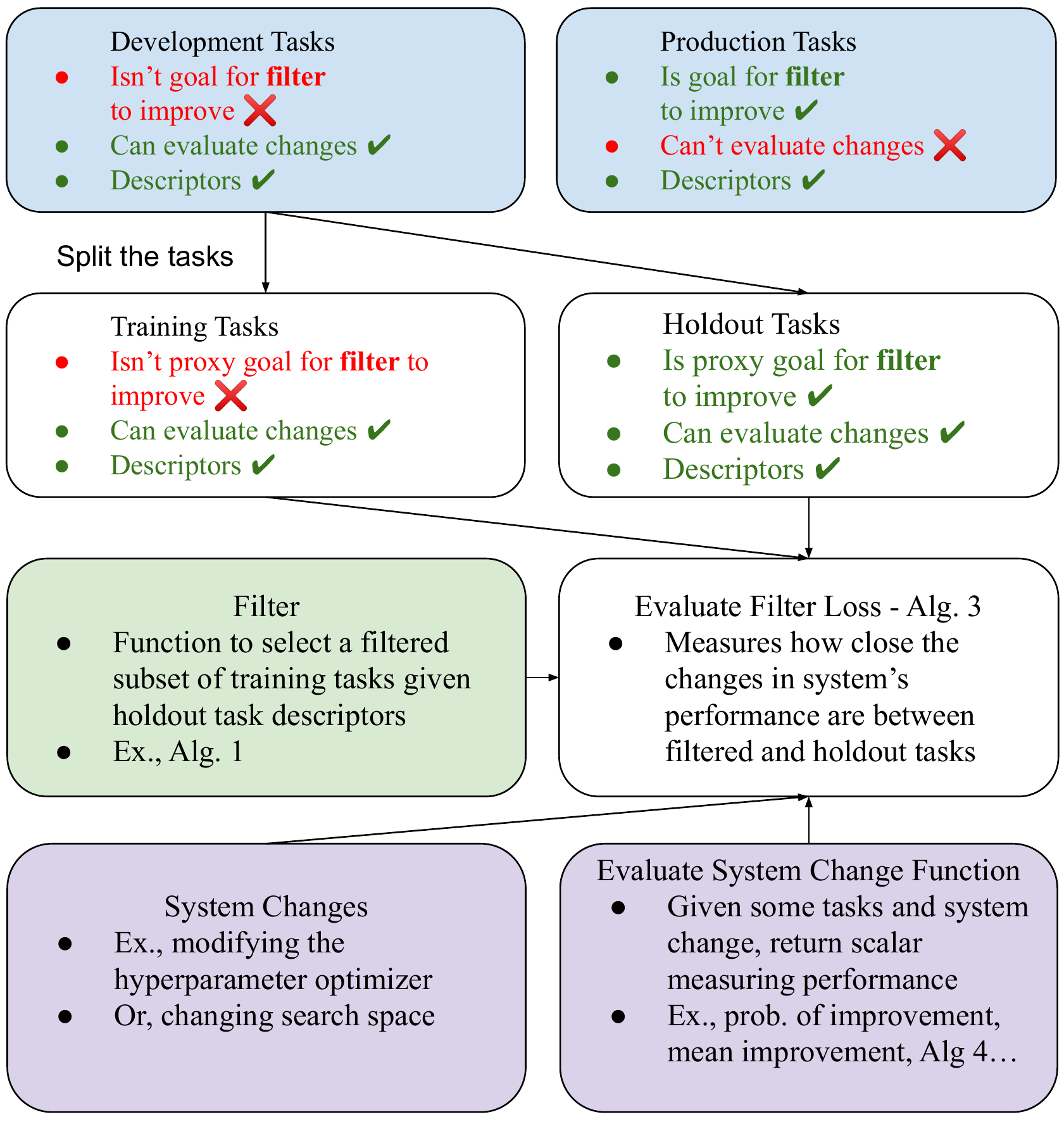}};
                \node[above=of img2, node distance=0cm, xshift=-.0cm, yshift=-1.2cm,font=\color{black}] {\footnotesize{{\color{red}Designing} Filters for Production Deployment Decisions}};
            \end{tikzpicture}
            \vspace{-0.015\textheight}
            \caption{
                \emph{Left:}
                How a filter is used to select relevant filtered, development tasks (given production task descriptors), which are evaluated on a system change for yes/no deployment decisions.
                \emph{Right:}
                The filter is tuned for effective deployment decisions, but we can not evaluate changes on production -- see Sec.~\ref{sec:service_provider_issue} -- so we select holdout tasks with no constraints to use as a proxy.
            }\label{fig:problem_summary}
            \vspace{-0.02\textheight}
        \end{figure}
    \section{Problem Setup} \label{sec:methods}
    \sectionSpaceAfter
        
        
        We propose a method allowing us to better assess AutoML system changes in \prod tasks by using their task descriptor to select relevant \dev tasks.
        More formally, we assume that when making a roll-out decision on a change, the service provider has descriptors for:
        (a) \dev tasks on the baseline setup.
        (b) \prod tasks on the baseline setup.
        (c) \dev tasks on the modified setup.

        Assessing AutoML system changes then boils down to the following steps -- see Fig.~\ref{fig:problem_summary}, left.
        (1) Filtering a subset of \dev tasks that more closely match \prod tasks.
        (2) Assessing the change on the filtered subset of \dev tasks.
        (3) Roll out the change to production if the modified setup improves over the baseline setup on the filtered subset of \dev tasks.
        In this paper, we propose a framework to design filters to perform step 1 of the above.

        Since we are limited in using \prod tasks for the purpose of this paper, we divide the \dev tasks into two sets as a proxy.
        From here on, \emph{train tasks} represent the role of \dev tasks and \emph{holdout tasks} are a proxy for \prod tasks.
        \Figure~\ref{fig:problem_summary} illustrates this setup.
        %
        In App. Sec.~\ref{sec:our_performance_measure} we describe a method to assess if a change to our AutoML system improves performance across a distribution of tasks by first measuring improvement on each task, then aggregating improvement measures across tasks.

    \sectionSpace
    \subsection{Filtering Tasks}\label{sec:filter_formulation}
    \sectionSpaceAfter
        The first step to assessing AutoML system changes is filtering \dev tasks that closely match the holdout tasks (a proxy for \prod tasks).
        Our filtering problem uses descriptor info from holdout tasks to select a filtered subset of the \dev tasks.
        This section presents different filtration strategies and then shows a method to evaluate the strategies. 
        

        \begin{wrapfigure}[7]{r}{0.51\textwidth}
            \vspace{-0.035\textheight}
            \begin{minipage}{1.0\linewidth}
            \begin{algorithm}[H]
                \caption{simFilter(length, simMetric)}\label{alg:similarity_filter}
                \begin{algorithmic}[1]
                    \State \textbf{def} filter(trainTasks, holdoutTasks):
                    \State \hspace{.05\textwidth}sims = simMetric(trainTasks,holdoutTasks)
                    \State \hspace{.05\textwidth}mostSimTasks =  argsort(sims)[:length]
                    \State \hspace{.05\textwidth}\textbf{return} trainTasks[mostSimTasks]
                    \State \textbf{return} filter
                    \Comment Note - this returns a function
                \end{algorithmic}
            \end{algorithm}
            \end{minipage}
        \end{wrapfigure}
        \subsubsection{Similarity Filters}\label{sec:similarity_filter}
            First, we design filters for a single holdout task, which we later aggregate into filters for multiple tasks.
            Intuitively, \emph{(sim)ilarity filter} methods take a way to measure the similarity between training tasks and a holdout task -- denoted a \emph{similarity metric} -- then returns the top $n$ tasks.
            \Algorithm~\ref{alg:similarity_filter} shows a skeleton for the proposed similarity filter parameterized by the number of returned tasks (\emph{filter length}) and the similarity metric.
            The returned item is a filter function taking train tasks and a holdout task, then returning a filtered subset of the train tasks.

        \begin{wrapfigure}[9]{r}{0.60\textwidth}
            \vspace{-0.015\textheight}
            \begin{minipage}{1.0\linewidth}
                \begin{algorithm}[H]
                    \caption{distanceSimMetric(trainTasks,holdoutTask)}\label{alg:feature_sim}
                    \begin{algorithmic}[1]
                        \State featureDistances = empty list
                        \For{trainTask in trainTasks}
                            \State \!\!\!\!\!$d$ \!\!=\!\! distance(trainTask.descriptor,\!\! holdoutTask.descriptor)\!\!\!
                            \State \!\!\!\!\!add $d$ to featureDistances
                        \EndFor
                        \State trainTaskSimilarities = $\nicefrac{1}{\textnormal{featureDistances}}$
                        \State \textbf{return} trainTaskSimilarities
                    \end{algorithmic}
                \end{algorithm}
            \end{minipage}
        \end{wrapfigure}
        \subsubsection{Similarity Metrics}\label{sec:similarity_measures}
            
            
            Our similarities metrics use the different types of info available to AutoML service providers in out setup, including:
            (a) task descriptors, (b) system descriptors.
            We also describe a similarity forming a heuristic performance bound.
            Sec.~\ref{sec:future_directions} lists other metrics that could be considered, but we did not use.
            
            \textbf{Task descriptor similarity:}
                For descriptors like the number of data points or features, we can reasonably rank train tasks in their euclidean distance from the holdout task, which we simply use as the similarity metric (Alg.~\ref{alg:feature_sim}).
                
            \textbf{Performance descriptor similarity:}
                For descriptors like the performance on different hyperparameter values, we can not simply compute a euclidean distance to the holdout task as with task descriptors.
                Instead we propose to use the following intuition: similar tasks have nearby qualities for hyperparameter values.
                Instead of evaluating train tasks on hyperparameter configurations used for the holdout tasks on the baseline setup, we estimate train tasks performance with a surrogate model.
                Then the correlation between the predicted quality and the actual quality from the holdout tasks is used as the similarity.
                Specific details are in App. Sec.~\ref{sec:app_similarity_measures}.

            \textbf{Oracle Similarity:}
                We would like a (relatively tight) upper bound on the possible performance of a similarity filter, to see how performant our filters are.
                Intuitively, we approximate this by constructing a filter using extra info about holdout tasks, which cannot be accessed via stored descriptors.
                Specifically, we re-run the holdout tasks with changes and simply compute correlations in task quality.
                See App. Sec.~\ref{sec:app_similarity_measures} for more details.
                This strategy is only a heuristic a upper bound and \emph{cannot be used on actual prod. tasks}, but proves useful in our experiments nonetheless.

        
        \sectionSpace
        \subsubsection{Constructing Filters for Multiple Holdout Tasks:}\label{sec:similarity_filter_multiple}
            Our prior sections looked at filters with one holdout task.
            However, we want filters for multiple holdout tasks because we have multiple tasks in prod.
            A simple approach for transforming a single holdout task filter is taking the union of the filtered tasks over every holdout task.
            However, this does not control the number of filtered tasks and can include rarely selected train tasks.
            We apply a filter for each holdout task and have them vote on the selected tasks.
            Once the votes are collected, we return the top $n$ tasks for a user-specified $n$.
            App. \Algorithm~\ref{alg:voting_filter} shows our voting method, and App. Sec.~\ref{sec:app_similarity_filter_multiple} discusses design choices.
    
        \sectionSpace
        \subsubsection{How to evaluate a filter with a loss}\label{sec:how_evaluate_filter}
            Now that we have a set of filters defined, we want to evaluate them.
            An ideal filter has the same result for changes on the filtered and holdout tasks.
            Hence, comparing the evaluation results with changes should inform us of filter strength.
            We show an example system change evaluating method in App. \Algorithm~\ref{alg:eval_change} (\emph{EvalSystemChange}).

            We propose a filter evaluation technique (\emph{EvalFilter}) in \Algorithm~\ref{alg:eval_filter} via a loss comparing system change results.
            \emph{EvalFilter} inputs a change to evaluate, a set of training \& holdout tasks, and a filter, then returns the scalar difference between the change's loss on filtered and holdout tasks.

            We measure the difference between the change's loss on the filtered and holdout tasks with a $\log$-loss because this is a common choice for comparing ($\log$)probs -- ex., as in logistic regression -- and our change's loss is a ($\log$)probability.
            The $\log$-loss is $t \log(y) + (1 - t) \log(1 - y)$, where $y, t$ are the probs. of improvement on the filtered and holdout tasks, respectively.
            \Figure~\ref{fig:filter_contrast_example_diffFilters} shows this loss with our oracle similarity and simple baselines, illustrating the range of values we should expect.
            %
            \begin{figure}[H]
                \vspace{-0.03\textheight}
                \begin{algorithm}[H]
                    \caption{EvalFilter(filter, trainTasks, holdoutTasks, change=(baselineSetup, modifiedSetup))}\label{alg:eval_filter}
                    \begin{algorithmic}[1]
                        \State filteredTasks = filter(trainTasks, holdoutTasks.descriptors)
                        \State $y$ = EvalSystemChange(filteredTasks, change)
                        \Comment The filtered task improvement probability
                        \State $t$ = EvalSystemChange(holdoutTasks, change)
                        \Comment The holdout task improvement probability
                        \State \textbf{return} logLoss = $t \log(y) + (1 - t) \log(1 - y)$
                        \Comment Can clip $y, t$ to prevent $\infty$
                    \end{algorithmic}
                \end{algorithm}
                \vspace{-0.04\textheight}
            \end{figure}
            

            \begin{figure}[t!]
                \centering
                \vspace{-0.05\textheight}
                \begin{tikzpicture}
                    \centering
                    \node (img){\includegraphics[trim={.75cm .75cm .61cm .5cm},clip, width=.4625\linewidth]{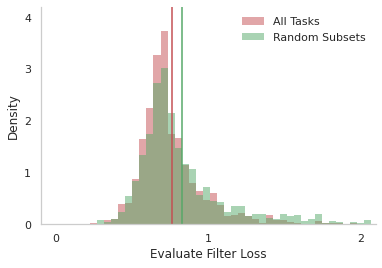}};
                    \node[left=of img, node distance=0cm, rotate=90, xshift=1.0cm, yshift=-.9cm, font=\color{black}] {\small{Density}};
                    \node[below=of img, node distance=0cm, xshift=3.5cm, yshift=1.25cm,font=\color{black}] {\small{Filter loss for various filters}};
                    
                    \node [right=of img, xshift=-1cm](img2){\includegraphics[trim={1.25cm .75cm .61cm 1.1cm},clip, width=.4675\linewidth]{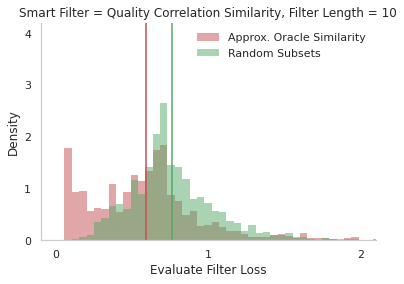}};
                \end{tikzpicture}
                \vspace{-0.0125\textheight}
                \caption{
                    We show two examples of contrasting filters with Alg.~\ref{alg:contrast_filter} using the loss from Alg.~\ref{alg:eval_filter}.
                    For each filter, we show the distribution of loss values from different samplings of train tasks and holdout tasks, along with a solid vertical line at the mean.
                }\label{fig:filter_contrast_example_diffFilters}
            \end{figure}
    \sectionSpace
    \section{Experiments} \label{sec:experiments}
    \sectionSpaceAfter
        First, we investigate the setup for our problem in Sec.~\ref{sec:filter_setup}.
        Next, we explore our filtering problem in Sec.~\ref{sec:filter_results}, showing that our filters improve over standard approaches.
        App. Sec.~\ref{sec:app_experiment} contains additional details for our experiments, including computational requirements.

        \sectionSpace
        \subsection{Experimental Setup}\label{sec:filter_setup}
        \sectionSpaceAfter
            To preface the filtering results in Sec.~\ref{sec:filter_results}, we first provide our setup, including:
            (a) changes made to our AutoML system.
            (b) the train and holdout task partitions.
            In our experiments we focus on binary classification with AUC as the quality, and say a system has performed better -- on a single run, for a single task -- if the quality measured is higher.
            
            \textbf{Changes made to the AutoML system for experiments:}\label{sec:configuration_results}
                We take a default setup and contrast with modified setups as in App. \Algorithm~\ref{alg:eval_change}.
                Our default uses a full, conditional search space for a fixed wall-clock time budget.
                We use a diverse range of modifications -- see App. Sec.~\ref{sec:app_configuration_results} -- including setups to assess changes to: 
                (1) search space, restricting to only a DNN.
                (2) hyperparameter optimizer by turning on transfer learning from \citep{golovin2017google}.
                (3) Vizier optimizer budget allowing $5\times$ more wall-clock time for hyperparameter queries.
                (4) the entire AutoML system by changing the underlying library used to implement the different learning algorithms.
                App. \Figure.~\ref{fig:config_example} displays the quality distribution for setups on various tasks, showing changes affect model quality in varied ways.
                Some tasks have positively correlated qualities, while some are anti-correlated, showing a simple similarity metric used in App. \Algorithm~\ref{alg:true_sim}.

            \textbf{Training and holdout task selection for experiments:}\label{sec:train_and_holdout_selection}
                Since we cannot use the \prod tasks for this paper, we divide the \dev tasks into two sets, \emph{train tasks} and \emph{holdout tasks}.
                We do this in two major ways for the results presented here:
                (a) The \dev tasks are assigned into either train or holdout at random, so the distribution of train and holdout tasks is similar.
                (b) The \dev tasks are partitioned into all OpenML tasks for train and all other tasks for holdout to simulate distribution shift.
                See App. Table~\ref{tab:training_details} for more details on specific tasks.

        \sectionSpace
        \subsection{Filtering Problem Results}\label{sec:filter_results}
\begin{figure}[t!]
                    \centering
                    \vspace{-0.05\textheight}
                    \begin{tikzpicture}
                        \centering
                        \node (img){\includegraphics[trim={1.0cm .75cm 1.2cm .75cm},clip, width=.33\linewidth]{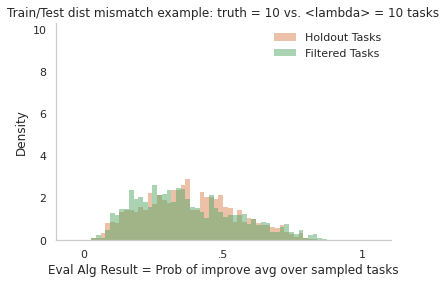}};
                        \node[left=of img, node distance=0cm, rotate=90, xshift=.8cm, yshift=-1.0cm, font=\color{black}] {\footnotesize{Density}};
                        \node[below=of img, node distance=0cm, xshift=4.35cm, yshift=1.25cm,font=\color{black}] {\footnotesize{Improvement probability for holdout and randomly filtered tasks}};
                        \node[above=of img, node distance=0cm, xshift=0cm, yshift=-.75cm,font=\color{black}] {\footnotesize{Holdout $\approx$ Train Tasks}};
                        \node[above=of img, node distance=0cm, xshift=0cm, yshift=-1.2cm,font=\color{black}] {\scriptsize{Change $=$ Default $\to$ DNN Only}};
                        
                        \node [right=of img, xshift=-1.5cm](img2){\includegraphics[trim={2.0cm .75cm 1.4cm .75cm},clip, width=.31\linewidth]{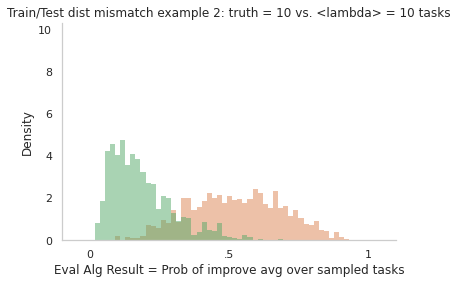}};
                        \node[above=of img2, node distance=0cm, xshift=0cm, yshift=-.75cm,font=\color{black}] {\footnotesize{{\color{red}Holdout $\neq$ Train Tasks}}};
                        \node[above=of img2, node distance=0cm, xshift=0cm, yshift=-1.2cm,font=\color{black}] {\scriptsize{Change $=$ Default $\to$ DNN Only}};
                        
                        \node [right=of img2, xshift=-1.75cm](img3){\includegraphics[trim={2.0cm .75cm 1.4cm .75cm},clip, width=.31\linewidth]{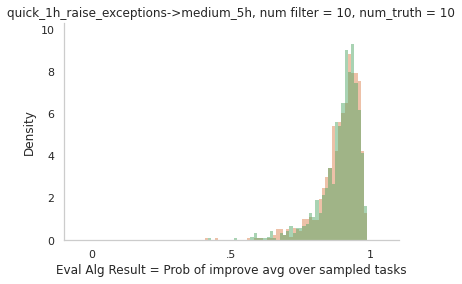}};
                        \node[above=of img3, node distance=0cm, xshift=0cm, yshift=-.75cm,font=\color{black}] {\footnotesize{Holdout $\approx$ Train Tasks}};
                        \node[above=of img3, node distance=0cm, xshift=0cm, yshift=-1.2cm,font=\color{black}] {\scriptsize{{\color{red}Change $=$ Default $\to$ $5\times$ compute}}};
                    \end{tikzpicture}
                    \vspace{-0.015\textheight}
                    \caption{
                        We look at the distribution of evaluate change results on a baseline random filter -- i.e., the improvement probability given the change.
                        \emph{Left:}
                        The resulting distribution when restricting the system search space to DNN only, and we sample our train and holdout tasks the same distribution.
                        We contrast this with other setups highlighted in {\color{red}red}.
                        \emph{Middle:}
                        The holdout tasks are sampled from a different distribution than train as specified in Section~\ref{sec:filter_setup}.
                        \emph{Right:}
                        We vary the system change to increase the compute budget by $5\times$.
                    }\label{fig:filter_eval_changes_and_mismatch}
                    \vspace{-0.02\textheight}
                \end{figure}

        \sectionSpaceAfter
            First, we compare different filters in Sec.~\ref{sec:contrast_filter_results}.
            Then, we use filters when holdout tasks are from a different distribution in Sec.~\ref{sec:larger_holdout_results}, demonstrating our proposed filters improve our ability to assess AutoML system change performance effects on holdout tasks with realistic distribution shifts.
            We also include investigations of design choices including evaluating AutoML system changes (Sec.~\ref{sec:eval_change_results}), filter strength (Sec.~\ref{sec:eval_filter_results}), and which changes filtration is useful for (\Figure~\ref{fig:filter_eval_changes_and_mismatch}).

            \subsubsection{Investigating design choices for evaluating changes to AutoML Systems}\label{sec:eval_change_results}
                \Algorithm~\ref{alg:eval_change} evaluates our system changes and has arguments of the evaluation tasks and system change (from a baseline setup to a modified setup).
                App. \Figure~\ref{fig:eval_change_over_changes} contrasts the improvement probability for different AutoML system changes and task sizes, showing how the improvement probability depends on the change we make to our system.
                Some changes -- like increasing the compute budget -- almost always improve performance, showing an example where the selected task does not affect the result. 
                
                We also display different numbers of tasks, showing how \Algorithm~\ref{alg:eval_change} behaves as we vary the filter and holdout sizes, verifying we can compare improvements on the differently sized filter and holdout as in later experiments (ex., \Figure~\ref{fig:analyze_filter_example_full} or \ref{fig:larger_holdout_sameDist}).
                App. \Figure~\ref{fig:eval_change_over_tasks} contrasts runs with a probability~estimate using one or multiple task evaluations, showing that more evaluations allow us to assess the improvement probability better.

            \subsubsection{Investigating how we evaluate filters}\label{sec:eval_filter_results}
                We now wield our ability to evaluate a change's loss in our AutoML system for evaluating filters.
                Alg.~\ref{alg:eval_filter} computes a loss for a filter given a set of train tasks, holdout tasks, and a system change with the difference between losses on the filtered and holdout tasks (for the change).
                This loss depends on the number of filtered and holdout tasks, how the train and holdout tasks differ, and the system change.
                As such, we vary these in our experiments.
                
                \Figure~\ref{fig:filter_eval_changes_and_mismatch}, left, evaluates the random task filter with the same train and holdout task distributions, showing the same results distributions with moderate variance.
                Here, filtering will only provide benefit if we have a limited number of tasks output by the filter.
                
                \Figure~\ref{fig:filter_eval_changes_and_mismatch}, middle, evaluates the random task filter while varying the train and holdout task distributions, showing that random selection performs poorly under mismatch.
                Here, we can benefit from a better filtering approach than the baseline strategy of random.
                
                \Figure~\ref{fig:filter_eval_changes_and_mismatch}, right, evaluates the random filter while varying the AutoML system change showing all filters perform equally well -- with near $0$ loss as in App. Fig.~\ref{fig:filter_eval_loss_dist} -- if a change always improves performance.
                Here, filtering will provide little to no benefit for any number of filtered tasks, because all tasks perform similarly.
                
                App. \Figure~\ref{fig:filter_eval_example} repeats these plots while varying the number of filtered and holdout tasks, showing we can compare the improvement probability distribution for differing task sizes.
            
            \subsubsection{Comparing filtering strategies on 1 holdout task}\label{sec:contrast_filter_results}
\begin{figure}[t!]
                    \centering
                    \vspace{-0.05\textheight}
                    \begin{tikzpicture}
                        \centering
                        \node (img){\includegraphics[trim={.75cm .75cm .5cm 1.1cm},clip, width=.48\linewidth]{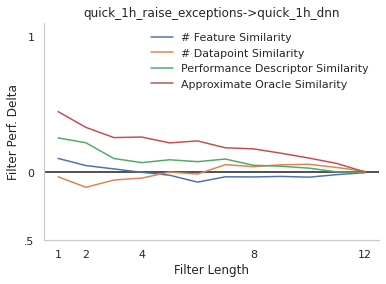}};
                        \node[left=of img, node distance=0cm, rotate=90, xshift=2.6cm, yshift=-1cm, font=\color{black}] {\small{Loss diff.~from random baseline}};
                        \node[below=of img, node distance=0cm, xshift=3.75cm, yshift=1.15cm,font=\color{black}] {\small{Number of tasks selected by filter}};
                        \node[above=of img, node distance=0cm, xshift=0cm, yshift=-1.2cm,font=\color{black}] {\small{Change $=$ Default $\to$ DNN Only}};
                        
                        \node [right=of img, xshift=-1cm](img2){\includegraphics[trim={1.25cm .75cm .5cm 1.1cm},clip, width=.465\linewidth]{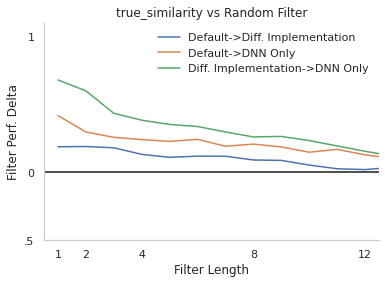}};
                        \node[above=of img2, node distance=0cm, xshift=0cm, yshift=-1.2cm,font=\color{black}] {\small{Approximate Oracle Similarity Filter}};
                    \end{tikzpicture}
                    \vspace{-0.03\textheight}
                    \caption{
                        We display the $5$-sample mean loss differences between filter strategies and a random baseline.
                        \emph{Left:} We show \textbf{filter strategies} in each {\color{blue}c}{\color{yellow}o}{\color{green}l}{\color{red}or} -- see App. Fig.~\ref{fig:analyze_filter_example_diffFilters} for other system changes.
                        \emph{Right:} We show the largest possible loss difference for \textbf{system changes} in each {\color{blue}c}{\color{yellow}o}{\color{green}l}{\color{red}or} with our heuristic upper bound  -- see App. Fig.~\ref{fig:analyze_filter_example_diffChanges} for other filters.
                    }\label{fig:analyze_filter_example_full}
                    \vspace{-0.02\textheight}
                \end{figure}

                %
                %
                \Figure~\ref{fig:analyze_filter_example_full} shows that various filtration strategies offer improvements in filter loss over a random filter baseline, even as we vary the filter lengths, similarities, and system changes.
                When we select all tasks, then all filters perform equally.
                Here, performance descriptor similarity is the best feasible filter for non-maximal tasks sizes, which we hypothesis is due to using extra info about the system, and not just the task.

                Also, App. \Figure~\ref{fig:analyze_filter_example_diffFilters}, \ref{fig:analyze_filter_example_diffChanges} and \ref{fig:analyze_filter_example_significance} show the best filtration will depend on the chosen system change by repeating this for more filtering strategies and system changes.
                Further, App. \Figure~\ref{fig:analyze_filter_example_significance} verifies that most filter loss differences pass a significance test.

            \subsubsection{Comparing filtering strategies on multiple holdout tasks:}\label{sec:larger_holdout_results}
\begin{figure}[t!]
                    \centering
                    \vspace{-0.05\textheight}
                    \begin{tikzpicture}
                        \centering
                        \node (img){\includegraphics[trim={.75cm 1.25cm .75cm 1.1cm},clip, width=.34\linewidth]{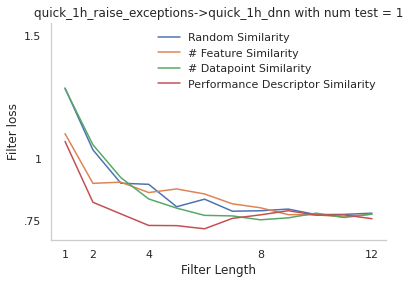}};
                        \node[left=of img, node distance=0cm, rotate=90, xshift=1.6cm, yshift=-.5cm, font=\color{black}] {\footnotesize{Holdout {\color{red}$\approx$} Train Tasks}};
                        \node[left=of img, node distance=0cm, rotate=90, xshift=1.4cm, yshift=-.9cm, font=\color{black}] {\small{Mean filter loss}};
                        \node[above=of img, node distance=0cm, xshift=0cm, yshift=-1.1cm,font=\color{black}] {\small{$1$ holdout task}};
                        
                        \node [right=of img, xshift=-1.4cm](img2){\includegraphics[trim={1.5cm 1.25cm .75cm 1.1cm},clip, width=.32\linewidth]{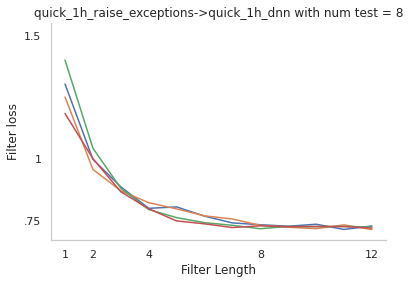}};
                        \node[above=of img2, node distance=0cm, xshift=0cm, yshift=-1.1cm,font=\color{black}] {\small{$8$ holdout tasks}};
                        
                        \node [right=of img2, xshift=-1.4cm](img3){\includegraphics[trim={1.5cm 1.25cm .75cm 1.1cm},clip, width=.32\linewidth]{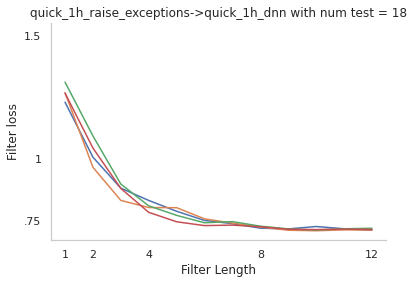}};
                        \node[above=of img3, node distance=0cm, xshift=0cm, yshift=-1.1cm,font=\color{black}] {\small{All $18$ holdout tasks}};
                        
                        \node [below=of img, yshift=1.1cm](img11){\includegraphics[trim={.75cm .75cm .75cm 1.1cm},clip, width=.34\linewidth]{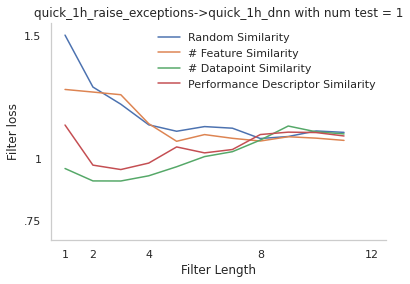}};
                        \node[left=of img11, node distance=0cm, rotate=90, xshift=1.6cm, yshift=-.5cm, font=\color{black}] {\footnotesize{Holdout {\color{red}$\neq$} Train Tasks}};
                        \node[left=of img11, node distance=0cm, rotate=90, xshift=1.5cm, yshift=-.9cm, font=\color{black}] {\small{Mean filter loss}};
                        
                        \node [right=of img11, xshift=-1.4cm](img21){\includegraphics[trim={1.5cm .75cm .75cm 1.1cm},clip, width=.32\linewidth]{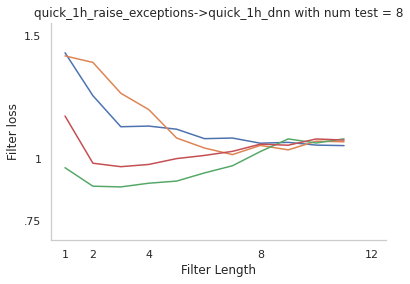}};
                        \node[below=of img21, node distance=0cm, xshift=.0cm, yshift=1.15cm,font=\color{black}] {\small{Number of tasks selected by filter}};
                        
                        \node [right=of img21, xshift=-1.4cm](img31){\includegraphics[trim={1.5cm .75cm .75cm 1.1cm},clip, width=.32\linewidth]{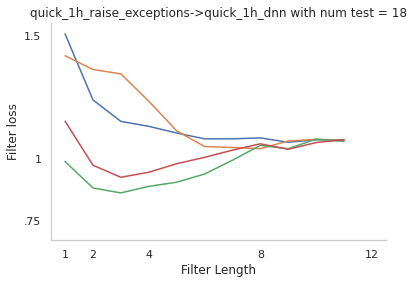}};
                    \end{tikzpicture}
                    \vspace{-0.03\textheight}
                    \caption{
                        We show different filters' absolute performance as we vary the number of holdout tasks, where we sample the holdout tasks from varying distributions described in Sec.~\ref{sec:filter_setup}.
                        \emph{Top:}
                        The random baseline always performs better with more tasks.
                        All methods perform equally well with many holdout tasks, as expected due to matching distributions.
                        \emph{Bottom:}
                        Our filters improve performance over the random baseline for any number of holdout tasks.
                    }\label{fig:larger_holdout_sameDist}
                    \vspace{-0.02\textheight}
                \end{figure}

                We construct multi-holdout-task filters using single-holdout-task filters (App. \Algorithm~\ref{alg:voting_filter}).
                We are particularly interested in holdout tasks with a distribution shift as a proxy for production.
                \Figure~\ref{fig:larger_holdout_sameDist} shows the absolute filter performance for various filters (each line), numbers of holdout tasks (left $\to$ right), and task distributions (top $\to$ bottom).
                Performance for all filters was similar at about $8$ selected tasks, so we display at most $12$.
                We use $30$ total tasks allocating the remaining $18$ to holdout.
                
                In \Figure~\ref{fig:larger_holdout_sameDist}, top -- where we sample the train and holdout tasks from the same distribution -- we do not see a benefit from filtering.
                The maximal filter size -- where all strategies are equal, and we use all tasks -- performs best because using more tasks is better when distributions match.
                
                Most importantly, in the bottom of \Figure~\ref{fig:larger_holdout_sameDist} -- where we sample the train and holdout tasks from different distributions -- filtering improves our ability to assess AutoML system change's affects on performance with larger holdout sizes.
                Notably, the best filters only uses $3$ tasks, with the \# data points then performance descriptor similarity.
                
            \sectionSpace
            
            \subsection{Takeaways}\label{sec:experiment_takeaways}
            \sectionSpaceAfter
                We see clear performance boosts from the filtering strategies when we have a limited number of holdout tasks, or the holdout task distribution is different from training.
                As such, we have the following takeaways for practitioners:
                \begin{enumerate}
                    \item \subsectionSpace If there are many \prod tasks, and the distribution differs from \dev, filtering helps -- \Figure~\ref{fig:larger_holdout_sameDist}, bottom.
                    Task (with $\#$ datapoints) and performance descriptor similarity performed best.
                    \item \subsectionSpace Using filtering, we saw similar performance using only $\approx 20\%$ of the tasks, providing a use-case for cheaper benchmarking with only $\approx 20\%$ the cost \emph{even with no \prod distribution shift} -- \Figure~\ref{fig:larger_holdout_sameDist}.
                    \item \subsectionSpace The best filter will depend on the system changes we use -- Sec.~\ref{sec:contrast_filter_results}.
                \end{enumerate}
            
            \subsection{Future Directions}\label{sec:future_directions}
            \sectionSpaceAfter
                
                Our primary goal is showing our filtering paradigm allows us to create simple filters that can select valuable tasks for assessing AutoML system changes -- not to construct complex filters.
                The next step is designing better filters because our simple baselines performed best.
                More complicated filters -- like KNNs -- could rank tasks using all available descriptor information.
                Alternatively, looking at more sophisticated metrics like problem complexity measures could be fruitful.
                
                We also found that simple task similarity metrics performed best, but these encode minimal information about the task.
                Investigating why these simple metrics performed robustly is interesting for future work.
                Also, we explored a range of changes to the system, hoping our results apply to other AutoML systems, but this should be experimentally verified.
                Finally, our filtering methods readily scale to magnitudes larger development task sets, so looking at how we should vary filtering in this regime may be necessary.

    \sectionSpace
    \section{Limitations \& Broader Impact}\label{sec:limitations}
    \sectionSpaceAfter
        \textbf{Limitations:}
        We used the improvement probability to assess system performance when evaluating system changes, but this might not accurately reflect a provider's risk tolerances.
        Users should select an appropriate scalar measure of system performance.
        Another limitation is that using improvement probability may not give helpful answers for changes that consistently improve performance, like increasing compute budget.
        There are other performance metric choices, like changes in the mean quality.
        Also, we explored a limited number of ways to measure task distribution mismatch -- there are various other strategies.

        \textbf{Broader Impact Statement:}
        While there are important impacts from using AutoML in general, we focus on a specific impact of our filtering formulation:
        Our method allows us to assess changes to our system using fewer development tasks, which could help reduce the environmental impact of benchmarking our system.
        For example, our best-performing filter used only $20\%$ of the available training tasks.
        However, this could create unintended biases by using fewer datasets.
        Also, our method could help service users who are specifically concerned with privacy.
        We should further investigate which groups most benefit from this.
        
    \sectionSpace 
    \section{Related work}\label{sec:related_work}
    \sectionSpace
        Our related work covers:
        AutoML broadly, AutoML systems we hope to apply our method on, then methods to evaluate those AutoML systems.
        Additional related work is in App. Sec.\ref{sec:app_related_work}.
        
        \textbf{AutoML:}
            \cite{hutter2019automated} contains an overview of key AutoML components -- including hyperparameter optimization~\cite{feurer_hyperparameter_2019}, meta-learning~\cite{vanschoren_meta_2019}, and neural architecture search~\cite{elsken_neural_2019}.
            
        \textbf{AutoML Systems:}
            We can apply our filters to arbitrary AutoML systems used for production.
            This includes various publicly available AutoML systems like Auto-WEKA~\cite{kotthoff_auto_2019}, Auto-MEKA~\cite{de2018automated}, Hyperopt-Sklearn~\cite{komer_hyperopt_2019}, Auto-sklearn~\cite{feurer_auto_2018}, Auto-Net~\cite{mendoza_towards_2019}, TPOT~\cite{olson_tpot_2019}, adaptive TPOT~\cite{evans2020adaptive}, the automatic statistician~\cite{steinruecken_automatic_2019}, AlphaD3m~\cite{drori2021alphad3m}, H20~\cite{ledell2020h2o}, SmartML~\cite{maher2019smartml}, ML-Plan~\cite{mohr2018ml}, Mosaic~\cite{rakotoarison2019automated}, RECIPE~\cite{de2017recipe}, Alpine Meadow~\cite{shang2109alpine}, ATM~\cite{swearingen2017atm}, Rafiki~\cite{wang2018rafiki}
            We focus on systems for multiple models, but some systems focus on DNNs including MetaQNN~\cite{baker2016designing}, Auto-Keras~\cite{jin2019auto}, and the various papers on Neural Architecture Search~\cite{zoph2016neural, pham2018efficient, maziarz2019evolutionary, adam2019understanding, jastrzkebski2018neural}.
            
        \textbf{Evaluating AutoML Systems:}
            There are various proposed benchmarks including HPO-B~\cite{arango2021hpo}, HPOBench~\cite{eggensperger2021hpobench}, AutoML Benchmark~\cite{gijsbers2019open}, and other varied studies~\cite{zoller2021benchmark, bischl2017openml, balaji2018benchmarking,  elshawi2019automated, he2021automl}.
            \cite{milutinovic2020evaluation} outline fundamental AutoML evaluation practices.
            Competitions are another method to provide common goals for AutoML systems, including the AutoML Challenges~\cite{guyon_analysis_2019}, the AutoDL challenge~\cite{liu2021winning} and NeurIPS BBO Challenge~\cite{turner2021bayesian}.

        \textbf{Filtering for AutoML:}
            There are fewer works on filtering strategies in AutoML.
            The closest work we are aware of in this vein is Oboe~\cite{yang2019oboe} using collaborative filtering on the matrix of qualities for many tasks for time-constrained HO.
    
    \sectionSpace
    \section{Conclusion} \label{sec:conclusion}
    \sectionSpaceAfter
        We were motivated by assessing if AutoML system changes – i.e., the search space or HPO – will improve the final output model's performance on a separate set of \prod tasks.
        But, we cannot run the system changes \prod tasks, so we assess them on \dev tasks.
        However, the set of \dev and \prod tasks differ, leading us to pursue changes improving \dev and not \prod
        We proposed the filtering problem, leveraging available descriptor info about holdout tasks to select useful \dev tasks.
        Then we proposed various filtration strategies, which we used in large-scale empirical studies.
        We showed that the filters improved our ability to assess if system changes improve performance on holdout tasks from different distributions than training, such as in \prod
        We hope this helps build the set of benchmarking strategies for more sophisticated and realistic setups, allowing people to better deploy AutoML systems.
        We also believe this provides a fruitful avenue of research in stronger filtration strategies, leveraging the broad body of work on task relationship learning.
        
    \section*{Acknowledgements}
        We would also like to thank Sagi Perel, and Luke Metz for feedback on this work and acknowledge the Python community ~\citep{van1995python, oliphant2007python} for developing the tools that enabled this work, including numpy 
        ~\citep{oliphant2006guide, van2011numpy, harris2020array}, Matplotlib~\citep{hunter2007matplotlib} and SciPy~\citep{jones2001scipy}.
      
    {\footnotesize
        \bibliography{references}

\begin{thebibliography}{81}
\providecommand{\natexlab}[1]{#1}
\providecommand{\url}[1]{\texttt{#1}}
\expandafter\ifx\csname urlstyle\endcsname\relax
  \providecommand{\doi}[1]{doi: #1}\else
  \providecommand{\doi}{doi: \begingroup \urlstyle{rm}\Url}\fi

\bibitem[Z{\"o}ller and Huber(2021)]{zoller2021benchmark}
Marc-Andr{\'e} Z{\"o}ller and Marco~F Huber.
\newblock Benchmark and survey of automated machine learning frameworks.
\newblock \emph{Journal of artificial intelligence research}, 70:\penalty0
  409--472, 2021.

\bibitem[Arango et~al.(2021)Arango, Jomaa, Wistuba, and
  Grabocka]{arango2021hpo}
Sebastian~Pineda Arango, Hadi~S Jomaa, Martin Wistuba, and Josif Grabocka.
\newblock Hpo-b: A large-scale reproducible benchmark for black-box hpo based
  on openml.
\newblock \emph{arXiv preprint arXiv:2106.06257}, 2021.

\bibitem[Eggensperger et~al.(2021)Eggensperger, M{\"u}ller, Mallik, Feurer,
  Sass, Klein, Awad, Lindauer, and Hutter]{eggensperger2021hpobench}
Katharina Eggensperger, Philipp M{\"u}ller, Neeratyoy Mallik, Matthias Feurer,
  Ren{\'e} Sass, Aaron Klein, Noor Awad, Marius Lindauer, and Frank Hutter.
\newblock Hpobench: A collection of reproducible multi-fidelity benchmark
  problems for hpo.
\newblock \emph{arXiv preprint arXiv:2109.06716}, 2021.

\bibitem[Bergstra et~al.(2011)Bergstra, Bardenet, Bengio, and
  K{\'e}gl]{bergstra2011algorithms}
James Bergstra, R{\'e}mi Bardenet, Yoshua Bengio, and Bal{\'a}zs K{\'e}gl.
\newblock Algorithms for hyper-parameter optimization.
\newblock \emph{Advances in neural information processing systems}, 24, 2011.

\bibitem[Hutter et~al.(2011)Hutter, Hoos, and
  Leyton-Brown]{hutter2011sequential}
Frank Hutter, Holger~H Hoos, and Kevin Leyton-Brown.
\newblock Sequential model-based optimization for general algorithm
  configuration.
\newblock In \emph{International conference on learning and intelligent
  optimization}, pages 507--523. Springer, 2011.

\bibitem[Snoek et~al.(2015)Snoek, Rippel, Swersky, Kiros, Satish, Sundaram,
  Patwary, Prabhat, and Adams]{snoek2015scalable}
Jasper Snoek, Oren Rippel, Kevin Swersky, Ryan Kiros, Nadathur Satish,
  Narayanan Sundaram, Mostofa Patwary, Mr~Prabhat, and Ryan Adams.
\newblock Scalable bayesian optimization using deep neural networks.
\newblock In \emph{International conference on machine learning}, pages
  2171--2180. PMLR, 2015.

\bibitem[Springenberg et~al.(2016)Springenberg, Klein, Falkner, and
  Hutter]{springenberg2016bayesian}
Jost~Tobias Springenberg, Aaron Klein, Stefan Falkner, and Frank Hutter.
\newblock Bayesian optimization with robust bayesian neural networks.
\newblock \emph{Advances in neural information processing systems}, 29, 2016.

\bibitem[Vanschoren et~al.(2013)Vanschoren, van Rijn, Bischl, and
  Torgo]{OpenML2013}
Joaquin Vanschoren, Jan~N. van Rijn, Bernd Bischl, and Luis Torgo.
\newblock Openml: Networked science in machine learning.
\newblock \emph{SIGKDD Explorations}, 15\penalty0 (2):\penalty0 49--60, 2013.
\newblock \doi{10.1145/2641190.2641198}.
\newblock URL \url{http://doi.acm.org/10.1145/2641190.2641198}.

\bibitem[Kohavi(1996)]{dataset_openml_adult}
Ron Kohavi.
\newblock Scaling up the accuracy of naive-bayes classifiers: A decision-tree
  hybrid.
\newblock In \emph{Proceedings of the Second International Conference on
  Knowledge Discovery and Data Mining}, KDD'96, page 202–207. AAAI Press,
  1996.

\bibitem[Moro et~al.(2014)Moro, Cortez, and
  Rita]{dataset_openml_bank_marketing}
Sérgio Moro, Paulo Cortez, and Paulo Rita.
\newblock A data-driven approach to predict the success of bank telemarketing.
\newblock \emph{Decision Support Systems}, 62:\penalty0 22--31, 2014.
\newblock ISSN 0167-9236.
\newblock \doi{https://doi.org/10.1016/j.dss.2014.03.001}.
\newblock URL
  \url{https://www.sciencedirect.com/science/article/pii/S016792361400061X}.

\bibitem[Candillier and Lemaire(2012)]{dataset_nomao}
Laurent Candillier and Vincent Lemaire.
\newblock Design and analysis of the nomao challenge - active learning in the
  real-world.
\newblock In \emph{Proceedings of the ALRA : Active Learning in Real-world
  Applications, Workshop ECML-PKDD 2012, Friday, September 28, 2012, Bristol,
  UK}, page to appear, 2012.

\bibitem[Guyon et~al.(2004)Guyon, Gunn, Ben-Hur, and Dror]{dataset_madelon}
Isabelle Guyon, Steve Gunn, Asa Ben-Hur, and Gideon Dror.
\newblock Result analysis of the nips 2003 feature selection challenge.
\newblock volume~17, 01 2004.

\bibitem[Johnson et~al.(2013)Johnson, Tateishi, and Hoan]{dataset_wilt}
Brian Johnson, Ryutaro Tateishi, and Nguyen Hoan.
\newblock A hybrid pansharpening approach and multiscale object-based image
  analysis for mapping diseased pine and oak trees.
\newblock \emph{International Journal of Remote Sensing}, 34:\penalty0
  6969--6982, 10 2013.
\newblock \doi{10.1080/01431161.2013.810825}.

\bibitem[Borkan et~al.(2019)Borkan, Dixon, Sorensen, Thain, and
  Vasserman]{dataset_civil_comments}
Daniel Borkan, Lucas Dixon, Jeffrey Sorensen, Nithum Thain, and Lucy Vasserman.
\newblock Nuanced metrics for measuring unintended bias with real data for text
  classification.
\newblock \emph{CoRR}, abs/1903.04561, 2019.
\newblock URL \url{http://arxiv.org/abs/1903.04561}.

\bibitem[Zhang et~al.(2015)Zhang, Zhao, and LeCun]{dataset_yelp_sentiment}
Xiang Zhang, Junbo Zhao, and Yann LeCun.
\newblock Character-level { {Convolutional Networks} } for { {Text
  Classification} }.
\newblock \emph{arXiv:1509.01626 [cs]}, September 2015.

\bibitem[{Diemert Eustache, Betlei Artem} et~al.(2018){Diemert Eustache, Betlei
  Artem}, Renaudin, and Massih-Reza]{dataset_criteo}
{Diemert Eustache, Betlei Artem}, Christophe Renaudin, and Amini Massih-Reza.
\newblock A large scale benchmark for uplift modeling.
\newblock In \emph{Proceedings of the AdKDD and TargetAd Workshop, KDD,
  London,United Kingdom, August, 20, 2018}. ACM, 2018.

\bibitem[Bischl et~al.(2017)Bischl, Casalicchio, Feurer, Hutter, Lang,
  Mantovani, van Rijn, and Vanschoren]{bischl2017openml}
Bernd Bischl, Giuseppe Casalicchio, Matthias Feurer, Frank Hutter, Michel Lang,
  Rafael~G Mantovani, Jan~N van Rijn, and Joaquin Vanschoren.
\newblock Openml benchmarking suites.
\newblock \emph{arXiv preprint arXiv:1708.03731}, 2017.

\bibitem[Balaji and Allen(2018)]{balaji2018benchmarking}
Adithya Balaji and Alexander Allen.
\newblock Benchmarking automatic machine learning frameworks.
\newblock \emph{arXiv preprint arXiv:1808.06492}, 2018.

\bibitem[Elshawi et~al.(2019)Elshawi, Maher, and Sakr]{elshawi2019automated}
Radwa Elshawi, Mohamed Maher, and Sherif Sakr.
\newblock Automated machine learning: State-of-the-art and open challenges.
\newblock \emph{arXiv preprint arXiv:1906.02287}, 2019.

\bibitem[Golovin et~al.(2017)Golovin, Solnik, Moitra, Kochanski, Karro, and
  Sculley]{golovin2017google}
Daniel Golovin, Benjamin Solnik, Subhodeep Moitra, Greg Kochanski, John Karro,
  and David Sculley.
\newblock Google vizier: A service for black-box optimization.
\newblock In \emph{Proceedings of the 23rd ACM SIGKDD international conference
  on knowledge discovery and data mining}, pages 1487--1495, 2017.

\bibitem[Hutter et~al.(2019{\natexlab{a}})Hutter, Kotthoff, and
  Vanschoren]{hutter2019automated}
Frank Hutter, Lars Kotthoff, and Joaquin Vanschoren, editors.
\newblock \emph{Automated Machine Learning - Methods, Systems, Challenges}.
\newblock Springer, 2019{\natexlab{a}}.

\bibitem[Feurer and Hutter(2019)]{feurer_hyperparameter_2019}
Matthias Feurer and Frank Hutter.
\newblock Hyperparameter optimization.
\newblock In  \citet{automl}, pages 3--38.

\bibitem[Vanschoren(2019)]{vanschoren_meta_2019}
Joaquin Vanschoren.
\newblock Meta-learning.
\newblock In  \citet{automl}, pages 39--68.

\bibitem[Elsken et~al.(2019)Elsken, Metzen, and Hutter]{elsken_neural_2019}
Thomas Elsken, Jan~Hendrik Metzen, and Frank Hutter.
\newblock Neural architecture search.
\newblock In  \citet{automl}, pages 69--86.

\bibitem[Kotthoff et~al.(2019)Kotthoff, Thornton, Hoos, Hutter, and
  Leyton-Brown]{kotthoff_auto_2019}
Lars Kotthoff, Chris Thornton, Holger~H. Hoos, Frank Hutter, and Kevin
  Leyton-Brown.
\newblock Auto-weka: Automatic model selection and hyperparameter optimization
  in weka.
\newblock In  \citet{automl}, pages 89--103.

\bibitem[de~S{\'a} et~al.(2018)de~S{\'a}, Freitas, and Pappa]{de2018automated}
Alex~GC de~S{\'a}, Alex~A Freitas, and Gisele~L Pappa.
\newblock Automated selection and configuration of multi-label classification
  algorithms with grammar-based genetic programming.
\newblock In \emph{International Conference on Parallel Problem Solving from
  Nature}, pages 308--320. Springer, 2018.

\bibitem[Komer et~al.(2019)Komer, Bergstra, and Eliasmith]{komer_hyperopt_2019}
Brent Komer, James Bergstra, and Chris Eliasmith.
\newblock Hyperopt-sklearn.
\newblock In  \citet{automl}, pages 105--121.

\bibitem[Feurer et~al.(2019)Feurer, Klein, Eggensperger, Springenberg, Blum,
  and Hutter]{feurer_auto_2018}
Matthias Feurer, Aaron Klein, Katharina Eggensperger, Jost~Tobias Springenberg,
  Manuel Blum, and Frank Hutter.
\newblock Auto-sklearn: Efficient and robust automated machine learning.
\newblock In  \citet{automl}, pages 123--143.

\bibitem[Mendoza et~al.(2019)Mendoza, Klein, Feurer, Springenberg, Urban,
  Burkart, Dippel, Lindauer, and Hutter]{mendoza_towards_2019}
Hector Mendoza, Aaron Klein, Matthias Feurer, Jost~Tobias Springenberg,
  Matthias Urban, Michael Burkart, Max Dippel, Marius Lindauer, and Frank
  Hutter.
\newblock Towards automatically-tuned deep neural networks.
\newblock In  \citet{automl}, pages 145--161.

\bibitem[Olson and Moore(2019)]{olson_tpot_2019}
Randal~S. Olson and Jason~H. Moore.
\newblock Tpot: A tree-based pipeline optimization tool for automating machine
  learning.
\newblock In  \citet{automl}, pages 163--173.

\bibitem[Evans et~al.(2020)Evans, Xue, and Zhang]{evans2020adaptive}
Benjamin Evans, Bing Xue, and Mengjie Zhang.
\newblock An adaptive and near parameter-free evolutionary computation approach
  towards true automation in automl.
\newblock In \emph{2020 IEEE Congress on Evolutionary Computation (CEC)}, pages
  1--8. IEEE, 2020.

\bibitem[Steinrucken et~al.(2019)Steinrucken, Smith, Janz, Lloyd, and
  Ghahramani]{steinruecken_automatic_2019}
Christian Steinrucken, Emma Smith, David Janz, James Lloyd, and Zoubin
  Ghahramani.
\newblock The automatic statistician.
\newblock In  \citet{automl}, pages 175--188.

\bibitem[Drori et~al.(2021)Drori, Krishnamurthy, Rampin, Lourenco, Ono, Cho,
  Silva, and Freire]{drori2021alphad3m}
Iddo Drori, Yamuna Krishnamurthy, Remi Rampin, Raoni de~Paula Lourenco,
  Jorge~Piazentin Ono, Kyunghyun Cho, Claudio Silva, and Juliana Freire.
\newblock Alphad3m: Machine learning pipeline synthesis.
\newblock \emph{arXiv preprint arXiv:2111.02508}, 2021.

\bibitem[LeDell and Poirier(2020)]{ledell2020h2o}
Erin LeDell and Sebastien Poirier.
\newblock H2o automl: Scalable automatic machine learning.
\newblock In \emph{Proceedings of the AutoML Workshop at ICML}, volume 2020,
  2020.

\bibitem[Maher and Sakr(2019)]{maher2019smartml}
Mohamed Maher and Sherif Sakr.
\newblock Smartml: A meta learning-based framework for automated selection and
  hyperparameter tuning for machine learning algorithms.
\newblock In \emph{EDBT: 22nd International Conference on Extending Database
  Technology}, 2019.

\bibitem[Mohr et~al.(2018)Mohr, Wever, and H{\"u}llermeier]{mohr2018ml}
Felix Mohr, Marcel Wever, and Eyke H{\"u}llermeier.
\newblock Ml-plan: Automated machine learning via hierarchical planning.
\newblock \emph{Machine Learning}, 107\penalty0 (8):\penalty0 1495--1515, 2018.

\bibitem[Rakotoarison et~al.(2019)Rakotoarison, Schoenauer, and
  Sebag]{rakotoarison2019automated}
Herilalaina Rakotoarison, Marc Schoenauer, and Mich{\`e}le Sebag.
\newblock Automated machine learning with monte-carlo tree search.
\newblock \emph{arXiv preprint arXiv:1906.00170}, 2019.

\bibitem[de~S{\'a} et~al.(2017)de~S{\'a}, Pinto, Oliveira, and
  Pappa]{de2017recipe}
Alex~GC de~S{\'a}, Walter Jos{\'e}~GS Pinto, Luiz Otavio~VB Oliveira, and
  Gisele~L Pappa.
\newblock Recipe: a grammar-based framework for automatically evolving
  classification pipelines.
\newblock In \emph{European Conference on Genetic Programming}, pages 246--261.
  Springer, 2017.

\bibitem[Shang et~al.()Shang, Zgraggen, and Kraska]{shang2109alpine}
Zeyuan Shang, Emanuel Zgraggen, and Tim Kraska.
\newblock Alpine meadow: A system for interactive automl.

\bibitem[Swearingen et~al.(2017)Swearingen, Drevo, Cyphers, Cuesta-Infante,
  Ross, and Veeramachaneni]{swearingen2017atm}
Thomas Swearingen, Will Drevo, Bennett Cyphers, Alfredo Cuesta-Infante, Arun
  Ross, and Kalyan Veeramachaneni.
\newblock Atm: A distributed, collaborative, scalable system for automated
  machine learning.
\newblock In \emph{2017 IEEE international conference on big data (big data)},
  pages 151--162. IEEE, 2017.

\bibitem[Wang et~al.(2018)Wang, Wang, Gao, Zhang, Chen, Ng, and
  Ooi]{wang2018rafiki}
Wei Wang, Sheng Wang, Jinyang Gao, Meihui Zhang, Gang Chen, Teck~Khim Ng, and
  Beng~Chin Ooi.
\newblock Rafiki: Machine learning as an analytics service system.
\newblock \emph{arXiv preprint arXiv:1804.06087}, 2018.

\bibitem[Baker et~al.(2016)Baker, Gupta, Naik, and Raskar]{baker2016designing}
Bowen Baker, Otkrist Gupta, Nikhil Naik, and Ramesh Raskar.
\newblock Designing neural network architectures using reinforcement learning.
\newblock \emph{arXiv preprint arXiv:1611.02167}, 2016.

\bibitem[Jin et~al.(2019)Jin, Song, and Hu]{jin2019auto}
Haifeng Jin, Qingquan Song, and Xia Hu.
\newblock Auto-keras: An efficient neural architecture search system.
\newblock In \emph{Proceedings of the 25th ACM SIGKDD international conference
  on knowledge discovery \& data mining}, pages 1946--1956, 2019.

\bibitem[Zoph and Le(2016)]{zoph2016neural}
Barret Zoph and Quoc~V Le.
\newblock Neural architecture search with reinforcement learning.
\newblock \emph{arXiv preprint arXiv:1611.01578}, 2016.

\bibitem[Pham et~al.(2018)Pham, Guan, Zoph, Le, and Dean]{pham2018efficient}
Hieu Pham, Melody Guan, Barret Zoph, Quoc Le, and Jeff Dean.
\newblock Efficient neural architecture search via parameters sharing.
\newblock In \emph{International conference on machine learning}, pages
  4095--4104. PMLR, 2018.

\bibitem[Maziarz et~al.(2019)Maziarz, Tan, Khorlin, Chang, Jastrz{\k{e}}bski,
  de~Laroussilhe, and Gesmundo]{maziarz2019evolutionary}
Krzysztof Maziarz, Mingxing Tan, Andrey Khorlin, Kuang-Yu~Samuel Chang,
  Stanis{\l}aw Jastrz{\k{e}}bski, Quentin de~Laroussilhe, and Andrea Gesmundo.
\newblock Evolutionary-neural hybrid agents for architecture search.
\newblock 2019.

\bibitem[Adam and Lorraine(2019)]{adam2019understanding}
George Adam and Jonathan Lorraine.
\newblock Understanding neural architecture search techniques.
\newblock \emph{arXiv preprint arXiv:1904.00438}, 2019.

\bibitem[Jastrz{\k{e}}bski et~al.(2018)Jastrz{\k{e}}bski, de~Laroussilhe, Tan,
  Ma, Houlsby, and Gesmundo]{jastrzkebski2018neural}
Stanis{\l}aw Jastrz{\k{e}}bski, Quentin de~Laroussilhe, Mingxing Tan, Xiao Ma,
  Neil Houlsby, and Andrea Gesmundo.
\newblock Neural architecture search over a graph search space.
\newblock \emph{arXiv preprint arXiv:1812.10666}, 2018.

\bibitem[Gijsbers et~al.(2019)Gijsbers, LeDell, Thomas, Poirier, Bischl, and
  Vanschoren]{gijsbers2019open}
Pieter Gijsbers, Erin LeDell, Janek Thomas, S{\'e}bastien Poirier, Bernd
  Bischl, and Joaquin Vanschoren.
\newblock An open source automl benchmark.
\newblock \emph{arXiv preprint arXiv:1907.00909}, 2019.

\bibitem[He et~al.(2021)He, Zhao, and Chu]{he2021automl}
Xin He, Kaiyong Zhao, and Xiaowen Chu.
\newblock Automl: A survey of the state-of-the-art.
\newblock \emph{Knowledge-Based Systems}, 212:\penalty0 106622, 2021.

\bibitem[Milutinovic et~al.(2020)Milutinovic, Schoenfeld, Martinez-Garcia, Ray,
  Shah, and Yan]{milutinovic2020evaluation}
Mitar Milutinovic, Brandon Schoenfeld, Diego Martinez-Garcia, Saswati Ray,
  Sujen Shah, and David Yan.
\newblock On evaluation of automl systems.
\newblock In \emph{Proceedings of the ICML Workshop on Automatic Machine
  Learning}, 2020.

\bibitem[Guyon et~al.(2019)Guyon, Sun-Hosoya, Boull{\'e}, Escalante, Escalera,
  Liu, Jajetic, Ray, Saeed, Sebag, Statnikov, Tu, and
  Viegas]{guyon_analysis_2019}
Isabelle Guyon, Lisheng Sun-Hosoya, Marc Boull{\'e}, Hugo~Jair Escalante,
  Sergio Escalera, Zhengying Liu, Damir Jajetic, Bisakha Ray, Mehreen Saeed,
  Mich{\`e}le Sebag, Alexander Statnikov, Wei-Wei Tu, and Evelyne Viegas.
\newblock Analysis of the automl challenge series 2015-2018.
\newblock In  \citet{automl}, pages 191--236.

\bibitem[Liu et~al.(2021)Liu, Pavao, Xu, Escalera, Ferreira, Guyon, Hong,
  Hutter, Ji, Junior, et~al.]{liu2021winning}
Zhengying Liu, Adrien Pavao, Zhen Xu, Sergio Escalera, Fabio Ferreira, Isabelle
  Guyon, Sirui Hong, Frank Hutter, Rongrong Ji, Julio CS~Jacques Junior, et~al.
\newblock Winning solutions and post-challenge analyses of the chalearn autodl
  challenge 2019.
\newblock \emph{IEEE Transactions on Pattern Analysis and Machine
  Intelligence}, 43\penalty0 (9):\penalty0 3108--3125, 2021.

\bibitem[Turner et~al.(2021)Turner, Eriksson, McCourt, Kiili, Laaksonen, Xu,
  and Guyon]{turner2021bayesian}
Ryan Turner, David Eriksson, Michael McCourt, Juha Kiili, Eero Laaksonen, Zhen
  Xu, and Isabelle Guyon.
\newblock Bayesian optimization is superior to random search for machine
  learning hyperparameter tuning: Analysis of the black-box optimization
  challenge 2020.
\newblock In \emph{NeurIPS 2020 Competition and Demonstration Track}, pages
  3--26. PMLR, 2021.

\bibitem[Yang et~al.(2019)Yang, Akimoto, Kim, and Udell]{yang2019oboe}
Chengrun Yang, Yuji Akimoto, Dae~Won Kim, and Madeleine Udell.
\newblock Oboe: Collaborative filtering for automl model selection.
\newblock In \emph{Proceedings of the 25th ACM SIGKDD International Conference
  on Knowledge Discovery \& Data Mining}, pages 1173--1183, 2019.

\bibitem[Van~Rossum and Drake~Jr(1995)]{van1995python}
Guido Van~Rossum and Fred~L Drake~Jr.
\newblock \emph{Python reference manual}.
\newblock Centrum voor Wiskunde en Informatica Amsterdam, 1995.

\bibitem[Oliphant(2007)]{oliphant2007python}
Travis~E Oliphant.
\newblock Python for scientific computing.
\newblock \emph{Computing in Science \& Engineering}, 9\penalty0 (3):\penalty0
  10--20, 2007.

\bibitem[Oliphant(2006)]{oliphant2006guide}
Travis~E Oliphant.
\newblock \emph{{A guide to NumPy}}, volume~1.
\newblock Trelgol Publishing USA, 2006.

\bibitem[Van Der~Walt et~al.(2011)Van Der~Walt, Colbert, and
  Varoquaux]{van2011numpy}
Stefan Van Der~Walt, S~Chris Colbert, and Gael Varoquaux.
\newblock {The NumPy array: A structure for efficient numerical computation}.
\newblock \emph{Computing in Science \& Engineering}, 13\penalty0 (2):\penalty0
  22--30, 2011.

\bibitem[Harris et~al.(2020)Harris, Millman, van~der Walt, Gommers, Virtanen,
  Cournapeau, Wieser, Taylor, Berg, Smith, et~al.]{harris2020array}
Charles~R Harris, K~Jarrod Millman, St{\'e}fan~J van~der Walt, Ralf Gommers,
  Pauli Virtanen, David Cournapeau, Eric Wieser, Julian Taylor, Sebastian Berg,
  Nathaniel~J Smith, et~al.
\newblock {Array programming with NumPy}.
\newblock \emph{Nature}, 585\penalty0 (7825):\penalty0 357--362, 2020.

\bibitem[Hunter(2007)]{hunter2007matplotlib}
John~D Hunter.
\newblock Matplotlib: A 2d graphics environment.
\newblock \emph{Computing in Science \& Engineering}, 9\penalty0 (3):\penalty0
  90--95, 2007.

\bibitem[Jones et~al.(2001)Jones, Oliphant, Peterson, et~al.]{jones2001scipy}
Eric Jones, Travis Oliphant, Pearu Peterson, et~al.
\newblock {SciPy: Open source scientific tools for Python}.
\newblock 2001.

\bibitem[Hutter et~al.(2019{\natexlab{b}})Hutter, Kotthoff, and
  Vanschoren]{automl}
Frank Hutter, Lars Kotthoff, and Joaquin Vanschoren, editors.
\newblock \emph{Automatic Machine Learning: Methods, Systems, Challenges}.
\newblock Springer, 2019{\natexlab{b}}.

\bibitem[Domke(2012)]{domke2012generic}
Justin Domke.
\newblock Generic methods for optimization-based modeling.
\newblock In \emph{Artificial Intelligence and Statistics}, pages 318--326.
  PMLR, 2012.

\bibitem[Maclaurin et~al.(2015)Maclaurin, Duvenaud, and
  Adams]{maclaurin2015gradient}
Dougal Maclaurin, David Duvenaud, and Ryan Adams.
\newblock Gradient-based hyperparameter optimization through reversible
  learning.
\newblock In \emph{International conference on machine learning}, pages
  2113--2122. PMLR, 2015.

\bibitem[Lorraine and Duvenaud(2018)]{lorraine2018stochastic}
Jonathan Lorraine and David Duvenaud.
\newblock Stochastic hyperparameter optimization through hypernetworks.
\newblock \emph{arXiv preprint arXiv:1802.09419}, 2018.

\bibitem[Mackay et~al.(2018)Mackay, Vicol, Lorraine, Duvenaud, and
  Grosse]{mackay2018self}
Matthew Mackay, Paul Vicol, Jonathan Lorraine, David Duvenaud, and Roger
  Grosse.
\newblock Self-tuning networks: Bilevel optimization of hyperparameters using
  structured best-response functions.
\newblock In \emph{International Conference on Learning Representations}, 2018.

\bibitem[Lorraine et~al.(2020)Lorraine, Vicol, and
  Duvenaud]{lorraine2020optimizing}
Jonathan Lorraine, Paul Vicol, and David Duvenaud.
\newblock Optimizing millions of hyperparameters by implicit differentiation.
\newblock In \emph{International Conference on Artificial Intelligence and
  Statistics}, pages 1540--1552. PMLR, 2020.

\bibitem[Lorraine et~al.(2022{\natexlab{a}})Lorraine, Acuna, Vicol, and
  Duvenaud]{lorraine2022complex}
Jonathan~P Lorraine, David Acuna, Paul Vicol, and David Duvenaud.
\newblock Complex momentum for optimization in games.
\newblock In \emph{International Conference on Artificial Intelligence and
  Statistics}, pages 7742--7765. PMLR, 2022{\natexlab{a}}.

\bibitem[Raghu et~al.(2021)Raghu, Lorraine, Kornblith, McDermott, and
  Duvenaud]{raghu2021meta}
Aniruddh Raghu, Jonathan Lorraine, Simon Kornblith, Matthew McDermott, and
  David~K Duvenaud.
\newblock Meta-learning to improve pre-training.
\newblock \emph{Advances in Neural Information Processing Systems},
  34:\penalty0 23231--23244, 2021.

\bibitem[Vicol et~al.(2022)Vicol, Lorraine, Pedregosa, Duvenaud, and
  Grosse]{vicol2022implicit}
Paul Vicol, Jonathan~P Lorraine, Fabian Pedregosa, David Duvenaud, and Roger~B
  Grosse.
\newblock On implicit bias in overparameterized bilevel optimization.
\newblock In \emph{International Conference on Machine Learning}, pages
  22234--22259. PMLR, 2022.

\bibitem[Lorraine et~al.(2022{\natexlab{b}})Lorraine, Vicol, Parker-Holder,
  Kachman, Metz, and Foerster]{lorraine2022lyapunov}
Jonathan Lorraine, Paul Vicol, Jack Parker-Holder, Tal Kachman, Luke Metz, and
  Jakob Foerster.
\newblock Lyapunov exponents for diversity in differentiable games.
\newblock In \emph{Proceedings of the 21st International Conference on
  Autonomous Agents and Multiagent Systems}, pages 842--852,
  2022{\natexlab{b}}.

\bibitem[Mo{\v{c}}kus(1975)]{movckus1975bayesian}
Jonas Mo{\v{c}}kus.
\newblock On bayesian methods for seeking the extremum.
\newblock In \emph{Optimization techniques IFIP technical conference}, pages
  400--404. Springer, 1975.

\bibitem[Snoek et~al.(2012)Snoek, Larochelle, and Adams]{snoek2012practical}
Jasper Snoek, Hugo Larochelle, and Ryan~P Adams.
\newblock Practical bayesian optimization of machine learning algorithms.
\newblock \emph{Advances in neural information processing systems}, 25, 2012.

\bibitem[Kandasamy et~al.(2020)Kandasamy, Vysyaraju, Neiswanger, Paria,
  Collins, Schneider, Poczos, and Xing]{kandasamy2020tuning}
Kirthevasan Kandasamy, Karun~Raju Vysyaraju, Willie Neiswanger, Biswajit Paria,
  Christopher~R Collins, Jeff Schneider, Barnabas Poczos, and Eric~P Xing.
\newblock Tuning hyperparameters without grad students: Scalable and robust
  bayesian optimisation with dragonfly.
\newblock \emph{J. Mach. Learn. Res.}, 21\penalty0 (81):\penalty0 1--27, 2020.

\bibitem[Wang et~al.(2021)Wang, Dahl, Swersky, Lee, Mariet, Nado, Gilmer,
  Snoek, and Ghahramani]{wang2021automatic}
Zi~Wang, George~E Dahl, Kevin Swersky, Chansoo Lee, Zelda Mariet, Zack Nado,
  Justin Gilmer, Jasper Snoek, and Zoubin Ghahramani.
\newblock Automatic prior selection for meta bayesian optimization with a case
  study on tuning deep neural network optimizers.
\newblock \emph{arXiv preprint arXiv:2109.08215}, 2021.

\bibitem[Feurer et~al.(2015)Feurer, Klein, Eggensperger, Springenberg, Blum,
  and Hutter]{feurer2015methods}
Matthias Feurer, Aaron Klein, Katharina Eggensperger, Jost Springenberg, Manuel
  Blum, and Frank Hutter.
\newblock Methods for improving bayesian optimization for automl.
\newblock In \emph{Proceedings of the International Conference on Machine
  Learning}, 2015.

\bibitem[Thornton et~al.(2013)Thornton, Hutter, Hoos, and
  Leyton-Brown]{thornton2013auto}
Chris Thornton, Frank Hutter, Holger~H Hoos, and Kevin Leyton-Brown.
\newblock Auto-weka: Combined selection and hyperparameter optimization of
  classification algorithms.
\newblock In \emph{Proceedings of the 19th ACM SIGKDD international conference
  on Knowledge discovery and data mining}, pages 847--855, 2013.

\bibitem[Friedman and Popescu(2008)]{friedman2008predictive}
Jerome~H Friedman and Bogdan~E Popescu.
\newblock Predictive learning via rule ensembles.
\newblock \emph{The annals of applied statistics}, 2\penalty0 (3):\penalty0
  916--954, 2008.

\bibitem[Cortes et~al.(2017)Cortes, Gonzalvo, Kuznetsov, Mohri, and
  Yang]{cortes2017adanet}
Corinna Cortes, Xavier Gonzalvo, Vitaly Kuznetsov, Mehryar Mohri, and Scott
  Yang.
\newblock Adanet: Adaptive structural learning of artificial neural networks.
\newblock In \emph{International conference on machine learning}, pages
  874--883. PMLR, 2017.

\bibitem[Song et~al.(2022)Song, Perel, Lee, Kochanski, and
  Golovin]{song2022open}
Xingyou Song, Sagi Perel, Chansoo Lee, Greg Kochanski, and Daniel Golovin.
\newblock Open source vizier: Distributed infrastructure and api for reliable
  and flexible blackbox optimization.
\newblock In \emph{First Conference on Automated Machine Learning (Main
  Track)}, 2022.

\end{thebibliography}
    }

    \newpage
    \appendix
    \onecolumn
    
    %
    \section*{Appendix}
        \section{Glossary}
            \begin{table*}[htbp]\caption{Glossary}
                \vspace{-0.03\textheight}
                \begin{center}
                    \begin{tabular}{c c}
                        \toprule
                        Search Space & The set of all possible hyper-parameters. \\
                        Task & The specification of the dataset and the learning problem.\\
                        (Prod.)uction tasks & Tasks observed in the production environment, presented by customers.\\
                        (Dev.)elopment tasks & Tasks used for development, constituting open-source datasets.\\
                        Filter & An algorithm for choosing a subset.\\
                        Train tasks & A subset of the \dev tasks used for designing a filter.\\
                        Holdout tasks & A subset of the \dev tasks used as a \\
                                    &representative of production tasks for evaluating a filter.\\
                        Filtered tasks & Subset of train tasks chosen by a filter.\\
                        Quality & A scalar quantity depicting the performance of a model on a task.\\
                        Baseline Setup & Some default AutoML setup.\\
                        Modified Setup & A non-default AutoML setup.\\
                        Change & A combination of a baseline and modified setup.\\
                        UB & Upper bound\\
                        DNN & Deep Neural Network\\
                        \bottomrule
                    \end{tabular}
                \end{center}
                \label{tab:TableOfNotation}
                \vspace{-0.01\textheight}
            \end{table*}

        \section{Additional Related work} \label{sec:app_related_work}
            \textbf{Hyperparameter optimization in AutoML:}
                    There are various scalable methods for HO using more information than quality evaluations~\cite{domke2012generic, maclaurin2015gradient, lorraine2018stochastic, mackay2018self, lorraine2020optimizing, lorraine2022complex, raghu2021meta}, which can have their own pitfalls\cite{vicol2022implicit, lorraine2022lyapunov}.
                    However, we focus on black-box HO like Bayesian optimization~\citep{snoek2015scalable, movckus1975bayesian, snoek2012practical, kandasamy2020tuning, wang2021automatic}.
                    \cite{feurer2015methods} focus on Bayesian optimization for AutoML.
                    We use Vizier~\cite{golovin2017google}, which supports various black-box optimizers including Bayesian Optimization.
        
        \section{Additional Background}\label{sec:app_background}
            \subsection{The Production AutoML System}\label{sec:app_our_automl_system_results}
                \subsubsection{The Search Space}\label{sec:AutoML_search_spaces}
                    Our AutoML system explores jointly algorithm selection and hyper-parameters~\cite{thornton2013auto}, modeled as a conditional search space, with a root parameter for the algorithm choice and conditional sub-parameters conditioned on the choice of the root node.
                    Our included learning algorithms are Feed-forward Neural Networks (DNNs), Random Forests, Gradient Boosted Decision Trees, Linear models, Sparse Linear Model with feature crosses (Cross) similar to RuleFit \cite{friedman2008predictive}, and ensembles of Linear and DNNs using AdaNet \cite{cortes2017adanet}.
                
                \subsubsection{The Hyperparameter Optimizer}\label{sec:AutoML_HO}
                    We use Vizier~\cite{golovin2017google, song2022open} to optimize our hyperparameters, which supports various user-specified parameters controlling the HPO.
                    Vizier supports various underlying HPO algorithms -- ex., random search, and Bayesian optimization.
                    We can specify parameters controlling the underlying HPO algorithms -- including exploration rate, whether to use transfer learning and how to apply early stopping.
                  
                \subsubsection{Assessing System Performance}\label{sec:AutoML_system_perf}
                    We focus on binary classification problems for simplicity here, but our work generalizes for an arbitrary scalar quality measure.
                    Specifically, we measure the AUC on the validation subset for the best model selected from the search space by our hyperparameter optimizer.
                    The system has performed better if we measure a higher quality on a single run for a single task.
                
        \section{The Filtering Problem}\label{sec:app_methodology}
                \textbf{Evaluating Changes to the AutoML System:}\label{sec:our_performance_measure}
                        To assess if some change to our AutoML system improves performance across a distribution of tasks, we first measure improvement on each task and then aggregate the improvement measures across tasks.
                        Note that quality measures between tasks are not directly comparable.
                        For example, even if two tasks use the same quality metric, such as AUC, the scale might be different.
                        In other words, an AUC of $0.92$ might be suitable for one task, while non-suitable for another task where $0.99$ AUC is possible.
                        To be able to compare quality across tasks, we measure improvement on a task with the probs.~ of the quality improving (\Algorithm~\ref{alg:eval_change}). 
                        To aggregate the improvement measure (i.e., probs.~ of improvement) across tasks, we could simply take the mean of the probabilities.
                        Instead, we use the mean of logits -- a simple transformation mapping the probs.~to $(-\infty, \infty)$ -- then going back to probs.~with the inverse-logit (\Algorithm~\ref{alg:eval_change}).
                        
                    \hspace{-0.05\textwidth}
                    \begin{minipage}[t!]{.99\linewidth}
                        \vspace{-0.01\textheight}
                        \begin{algorithm}[H]
                            \caption{EvalSystemChange(tasks, change=(baselineSetup, modifiedSetup))}\label{alg:eval_change}
                            \begin{algorithmic}[1]
                                \For{task in tasks}
                                \Comment compute every taskLoss for each task
                                    \State get baselineQualities from baselineSetup
                                    \Comment An array of qualities -- ex., from a database
                                    \State get modifiedQualities for each run with modifiedSetup
                                    \State probImproved = fraction of pairings with modifiedQuality > baselineQuality
                                    \State taskLoss = logit(probImproved)
                                    \Comment Loss is log-odds of improvement probability with change
                                \EndFor
                                \State \textbf{return} expit(meanTaskLoss)
                                \Comment Aggregate with mean, and take inverse logit for a probability
                            \end{algorithmic}
                        \end{algorithm}
                        \vspace{-0.01\textheight}
                    \end{minipage}
                        
                \subsection{How Filters work}\label{sec:app_how_filter_work}
                    \hphantom{a}
                    \vspace{-0.025\textheight}
                    \begin{algorithm}[H]
                        \caption{abstractFilter(trainTasks, holdoutTaskDescriptors)}\label{alg:example_filter}
                        \begin{algorithmic}[1]
                            \State Construct a list L containing elements of trainTasks, given holdoutTaskDescriptors
                            \State \textbf{return} L
                        \end{algorithmic}
                    \end{algorithm}
                
                \subsection{How to Compare Filters}\label{sec:app_how_compare_filter}
                    \hphantom{a}
                    \vspace{-0.025\textheight}
                    \begin{algorithm}[H]
                        \caption{contrastFilters(newFilter, baselineFilter, baselineSetup, modifiedSetup, taskPartitions)}\label{alg:contrast_filter}
                        \begin{algorithmic}[1]
                            \State Initialize loss storage arrays newFilterLosses, baselineFilterLosses
                            \For{each trainTasks, holdoutTasks in taskPartitions}
                                \State newFilterLoss = EvalFilter(newFilter, trainTasks, holdoutTasks, baselineSetup, modifiedSetup)
                                \State baselineFilterLoss = EvalFilter(newFilter, trainTasks, holdoutTasks, baselineSetup, modifiedSetup)
                                \State Store newFilterLoss and baselineFilterLoss
                            \EndFor
                            \State Compute summary of loss distributions -- ex., the difference in the mean losses, or a significance test to see if newFilter's loss is lower, significantly.
                            \State \textbf{return} filter loss distribution summary
                        \end{algorithmic}
                    \end{algorithm}
                    
                \subsection{Similarity Metrics}\label{sec:app_similarity_measures}
                    \hphantom{a}

                    
                    \textbf{Performance descriptor similarities}:
                        \Algorithm~\ref{alg:surrogate_sim} uses a surrogate model to estimate the performance of the hyperparameter configurations on a train task by taking the hyperparameter values as input features.
                        The intuition is that similar tasks perform similarity on the same hyperparameter values.
                        Instead of actually evaluating the train tasks on all hyperparameter configuration used for the holdout tasks on the baseline setup we estimate training performance using a surrogate model.
                        
                        The surrogate model is trained for each train task on past runs on all baseline setups.
                        Then the trained surrogate model are used to predict what performance metric will be for the hyperparameter values that were applied to holdout task tasks on the baseline setup.
                        These predictions give an idea of how the train tasks would have performed on the set of hyperparameter values used for the holdout tasks.
                        The correlation between the predicted quality/objective metric and the actual metrics from the holdout tasks are computed and used as a similarity measure between the train tasks and the holdout task.
                        We outline this method in \Algorithm~\ref{alg:surrogate_sim}.
                        \vspace{-0.015\textheight}
                        \begin{algorithm}[H]
                            \caption{performanceDescriptorSimilarity(trainTasks, holdoutTask)}\label{alg:surrogate_sim}
                            \begin{algorithmic}[1]
                                \State trainTaskSimilarities = empty list
                                \State Train surrogate models for each task in trainTasks.
                                \State holdoutTaskHparamConfigs, holdoutTaskQuality = fetch past runs for holdoutTask for baseline setup
                                \For{trainTask in trainTasks}
                                    \State surrogateModel = Fetch surrogate model for trainTask
                                    \State estimatedQuality = surrogateModel(holdoutTaskHparamConfigs)
                                    \State similarity = Compute a correlation coefficient for holdoutTaskQuality and estimatedQuality
                                    \State store similarity for task in trainTaskSimilarities
                                \EndFor
                                \State \textbf{return} trainTaskSimilarities
                            \end{algorithmic}
                        \end{algorithm}
                        \vspace{-0.015\textheight}
                        
                    \textbf{Approximate oracle similarity:}
                        When developing our filters, it would be desirable to have a (relatively tight) upper-bound on the possible performance given our limited access to info about the holdout tasks.
                        For example, once we approach this performance then we know it is not worth attempting to design better filters.
                        This also allows us to better gauge the magnitude of improvement over a random task selection baseline.
                        
                        We propose a heuristic to approximate an upper-bound on the filter performance using limited descriptors about holdout data, with a filter which uses the unrestricted descriptors about holdout data.
                        This setup allows re-running the holdout data with AutoML system changes.
                        If we can access this info about holdout tasks, then we wouldn't need to filter.
                        We would simply evaluate our changes on holdout tasks as is standard practice.
                        Nonetheless, this is useful to bound possible filter performance.
                        
                        Specifically, in App. \Algorithm~\ref{alg:true_sim} we compute the similarity by simply computing the correlation in qualities between a train task and test task -- sampled over different setups to our AutoML system.
                        The choice of setups to sample over, and the correlation coefficient to compute are user-specified.
                        This is heuristic with no strong guarantees, but proves useful in our experiments.
                        Experiments for App. \Algorithm~\ref{alg:true_sim} are in Sec.~\ref{sec:true_sim_results}.
                        \vspace{-0.015\textheight}
                        \begin{algorithm}[H]
                            \caption{oracleSimilarity(trainTasks, holdoutTask)}\label{alg:true_sim}
                            \begin{algorithmic}[1]
                                \State trainTaskSimilarities = empty list
                                \For{trainTask in trainTasks}
                                    \State trainQualities, holdoutQualities = empty lists
                                    \For{setup in systemSetups}
                                        \State store trainQuality = Run the train task with the setup
                                        \State store holdoutQuality = Run the holdout task with the setup
                                    \EndFor
                                    \State similarity = compute a correlation coefficient for trainQualities and holdoutQualities
                                    \State store similarity for task in trainTaskSimilarities
                                \EndFor
                                \State \textbf{return} trainTaskSimilarities
                            \end{algorithmic}
                        \end{algorithm}
                        \vspace{-0.015\textheight}
                        
                \subsubsection{Constructing Filters for Multiple Holdout Tasks}\label{sec:app_similarity_filter_multiple}
                    For \Algorithm~\ref{alg:voting_filter}, the number of filtered tasks for the inner, single-task filter and the outer multi-task filter can be different, but we keep them the same for simplicity in our experiments.
                    Also, each filtered task receives a single vote of equal weight.
                    We can use more complicated strategies like weighted votes -- ex., based on similarity scores -- but these were unnecessary for our experiments.
                    \vspace{-0.015\textheight}
                    \begin{algorithm}[H]
                        \caption{VotingFilter(trainTasks, testTasks, filterLength, innerFilter)}\label{alg:voting_filter}
                        \begin{algorithmic}[1]
                            \State initialize voteCounts = dictionary of how many votes each train task received
                            \For{testTask in testTasks}
                                \State filteredTasks = innerFilter(trainTasks, testTask)
                                \For{filteredTask in filteredTasks}
                                    \State voteCounts[filteredTask] += 1
                                \EndFor
                            \EndFor
                            \State \textbf{return} the top filterLength tasks with the most votes in voteCounts
                        \end{algorithmic}
                    \end{algorithm}
                    \vspace{-0.025\textheight}
                    
                \subsubsection{How to compare filters:}\label{sec:app_compare_filters}
                    Once we can evaluate a filter's loss, we use this to compare different filters.
                    App. \Algorithm~\ref{alg:contrast_filter} shows a skeleton for our method of comparing filters, with relevant experiments in Sec.~\ref{sec:contrast_filter_results}. 
                    We look at two simple strategies:
                    First, we compute the cross-entropy loss for each filter -- i.e., the filter log-loss averaged over different train and test task partitions -- and report the difference in cross-entropies as the magnitude of improvement between filtering strategies (\Figure~\ref{fig:analyze_filter_example_full}).
                    Second, we use a significance test on the difference in cross-entropy for the filters (\Figure~\ref{fig:analyze_filter_example_significance}).
        \section{Experiments}\label{sec:app_experiment}
            
            \subsection{Compute Requirements} \label{sec:compute_requirements}
                The total compute requirement is equivalent to running $60k$ CPUs for $1$ day, consuming $10$ PiB of memory on average.
            
            \vspace{-0.025\textheight}
            \begin{table*}[htbp]\caption{Training details}
                \vspace{-0.025\textheight}
                \begin{center}
                    \begin{tabular}{l l}
                        \toprule
                        Datasets & OpenML~\cite{OpenML2013} (Phishing Websites, Adult~\cite{dataset_openml_adult}, \\&Bank Marketing~\cite{dataset_openml_bank_marketing}, Churn, Electricity, Kr-vs-kp, Numerai28.6,\\& Sick, Spambase, Nomao~\cite{dataset_nomao}, Jm1, Internet Advertisements, Phoneme,\\& Madelon~\cite{dataset_madelon}, Bioresponse, Wilt~\cite{dataset_wilt}), Civil Comments~\cite{dataset_civil_comments}, Imdb, \\& Census, Yelp sentiment~\cite{dataset_yelp_sentiment}, Criteo~\cite{dataset_criteo} \\
                        Search Space & Random Forest, Linear, Deep Neural Networks, Gradient Boosted \\&Decision Trees, AdaNet, Linear Feature Cross \\
                        Data Splits & 80-10-10 split, uniformly at random.\\
                        \bottomrule
                    \end{tabular}
                \end{center}
                \label{tab:training_details}
            \end{table*}
            \vspace{-0.025\textheight}

            \subsection{Experimental Setup}\label{sec:app_filter_setup}
                
                \subsubsection{Changes made to the AutoML system for experiments}\label{sec:app_configuration_results}
                    \hphantom{a}
                    \begin{figure}[H]
                        \centering
                        \vspace{-0.02\textheight}
                        \begin{tikzpicture}
                            \centering
                            \node (img){\includegraphics[trim={.75cm .75cm .0cm .0cm},clip, width=.45\linewidth]{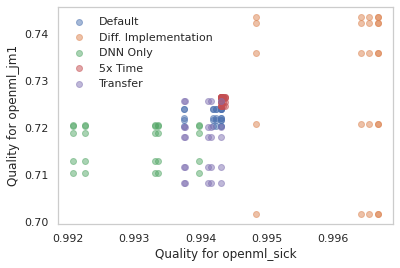}};
                            \node[left=of img, node distance=0cm, rotate=90, xshift=2.2cm, yshift=-.85cm, font=\color{black}] {\small{Quality for OpenML \texttt{jm1} task}};
                            \node[below=of img, node distance=0cm, xshift=.5cm, yshift=1.1cm,font=\color{black}] {\small{Quality for OpenML \texttt{sick} task}};
                            
                            \node [right=of img, xshift=-.35cm](img2){\includegraphics[trim={.75cm .75cm .0cm .0cm},clip, width=.45\linewidth]{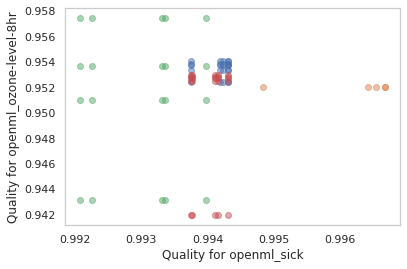}};
                            \node[left=of img2, node distance=0cm, rotate=90, xshift=2.25cm, yshift=-.85cm, font=\color{black}] {\small{Quality for OpenML \texttt{ozone} task}};
                            \node[below=of img2, node distance=0cm, xshift=-.0cm, yshift=1.1cm,font=\color{black}] {\small{Quality for OpenML \texttt{sick} task}};
                        \end{tikzpicture}
                        \vspace{-0.025\textheight}
                        \caption{
                            We show of the distribution of qualities for our setups on various pairs of OpenML tasks.
                            Each setups qualities are shown in a different color.
                            For each task and setup, we have multiple observations over re-runs, so we plot every pairing of the tasks qualities for each setup.
                        }\label{fig:config_example}
                        \vspace{-0.025\textheight}
                    \end{figure}
                    
                \subsubsection{Approximate Oracle Similarity Results}\label{sec:true_sim_results}
                    
\begin{figure}[H]
                        \vspace{-0.015\textheight}
                        \centering
                        \begin{tikzpicture}
                            \centering
                            \node (img){\includegraphics[trim={.0cm .0cm .0cm .75cm},clip, width=.5\linewidth]{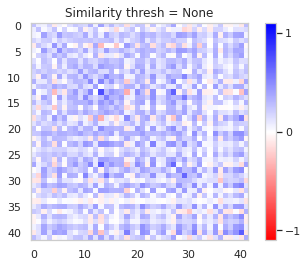}};
            
                            \node[left=of img, node distance=0cm, rotate=90, xshift=1.5cm, yshift=-1cm, font=\color{black}] {Task Index};
            
                            \node[below=of img, node distance=0cm, xshift=-.0cm, yshift=1.3cm,font=\color{black}] {Task Index};
                        \end{tikzpicture}
                        \vspace{-0.015\textheight}
                        \caption{
                            We show the matrix of approximate oracle similarities where each row and column correspond to a single task.
                        }\label{fig:true_sim_matrix}
                        \vspace{-0.015\textheight}
                    \end{figure}

                    
                    \Figure~\ref{fig:true_sim_matrix} display the matrix of similarities from \Algorithm~\ref{alg:true_sim}, computed using the system setups from Sec.~\ref{sec:app_configuration_results}, showing non-trivial task relationships.
                    
            \subsection{Filtering Problem Results}\label{sec:app_filter_results}

                
                \subsubsection{Investigating design choices for evaluating changes to AutoML Systems}\label{sec:app_eval_change_results}
                    
                    \hphantom{a}
                    \begin{figure}[H]
                        \centering
                        \vspace{-0.015\textheight}
                        \begin{tikzpicture}
                            \centering
                            \node (img){\includegraphics[trim={.75cm .75cm 1.1cm 1.0cm},clip, width=.46\linewidth]{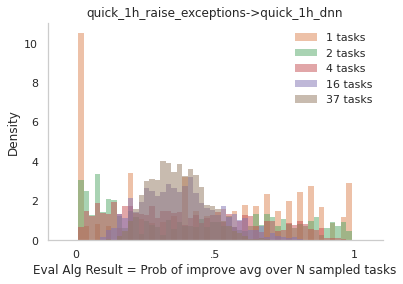}};
                            \node[left=of img, node distance=0cm, rotate=90, xshift=1.05cm, yshift=-.95cm, font=\color{black}] {\small{Density}};
                            \node[below=of img, node distance=0cm, xshift=3.75cm, yshift=1.25cm,font=\color{black}] {\small{Improvement probability for $n$ tasks}};
                            \node[above=of img, node distance=0cm, xshift=0cm, yshift=-1.25cm,font=\color{black}] {\small{Change $=$ Default $\to$ DNN Only}};
                            
                            \node [right=of img, xshift=-1cm](img2){\includegraphics[trim={.75cm .75cm 1.1cm 1.0cm},clip, width=.46\linewidth]{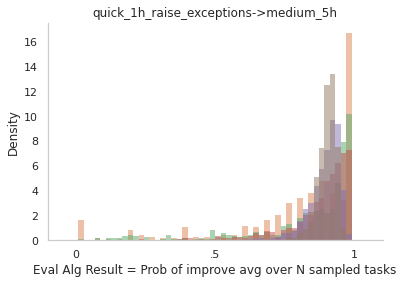}};
                            \node[above=of img2, node distance=0cm, xshift=0cm, yshift=-1.25cm,font=\color{black}] {\small{Change $=$ Default $\to$ $5\times$ compute}};
                        \end{tikzpicture}
                        \vspace{-0.015\textheight}
                        \caption{
                            We contrast runs of \emph{EvalSystemChange} (\Algorithm~\ref{alg:eval_change}) showing the improvement probability for different AutoML system changes as we vary the number of tasks.
                        }\label{fig:eval_change_over_changes}
                        \vspace{-0.015\textheight}
                    \end{figure}
                    
                    \begin{figure}[H]
                        \centering
                        \vspace{-0.015\textheight}
                        \begin{tikzpicture}
                            \centering
                            \node (img){\includegraphics[trim={.75cm .75cm .25cm .75cm},clip, width=.46\linewidth]{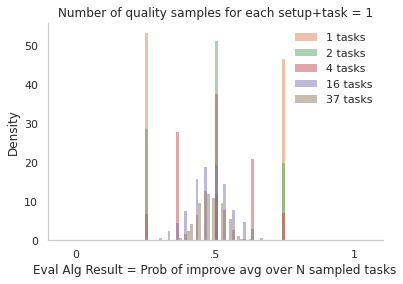}};
                            \node[left=of img, node distance=0cm, rotate=90, xshift=1.5cm, yshift=-.75cm, font=\color{black}] {Density};
                            \node[below=of img, node distance=0cm, xshift=3.75cm, yshift=1.25cm,font=\color{black}] {Improvement Prob. for $n$ tasks};
                            \node[above=of img, node distance=0cm, xshift=0cm, yshift=-1.4cm,font=\color{black}] {$1$ quality evaluation};
                            
                            \node [right=of img, xshift=-1cm](img2){\includegraphics[trim={.75cm .75cm .25cm .75cm},clip, width=.46\linewidth]{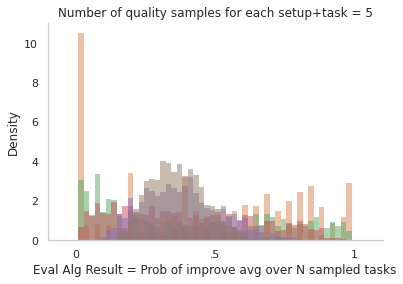}};
                            \node[above=of img2, node distance=0cm, xshift=0cm, yshift=-1.4cm,font=\color{black}] {$5$ quality evaluations};
                        \end{tikzpicture}
                        \vspace{-0.015\textheight}
                        \caption{
                            We contrast runs of \Algorithm~\ref{alg:eval_change} with a probability~estimate using a single task evaluation, with a probability~estimate using multiple task evaluations.
                        }\label{fig:eval_change_over_tasks}
                        \vspace{-0.015\textheight}
                    \end{figure}

                    \begin{figure}[H]
                        \centering
                        \vspace{-0.025\textheight}
                        \begin{tikzpicture}
                            \centering
                            \node (img){\includegraphics[trim={.75cm .75cm .25cm .75cm},clip, width=.34\linewidth]{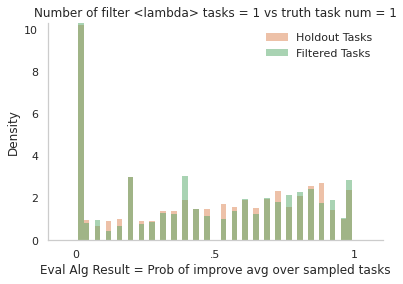}};
                            \node[left=of img, node distance=0cm, rotate=90, xshift=1.0cm, yshift=-1.0cm, font=\color{black}] {Density};
                            \node[above=of img, node distance=0cm, xshift=0cm, yshift=-1.2cm,font=\color{black}] {\small{$1$ filtered task, $1$ holdout task}};
                            
                            \node [right=of img, xshift=-1.2cm](img2){\includegraphics[trim={1.5cm .75cm .25cm .75cm},clip, width=.31\linewidth]{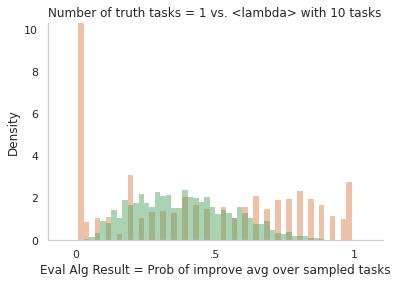}};
                            \node[below=of img2, node distance=0cm, xshift=-.0cm, yshift=1.15cm,font=\color{black}] {Improvement Prob. for Holdout and Filtered Tasks};
                            \node[above=of img2, node distance=0cm, xshift=0cm, yshift=-1.2cm,font=\color{black}] {\small{$10$ filtered tasks, $1$ holdout task}};
                            
                            \node [right=of img2, xshift=-1.2cm](img3){\includegraphics[trim={1.5cm .75cm .25cm .75cm},clip, width=.31\linewidth]{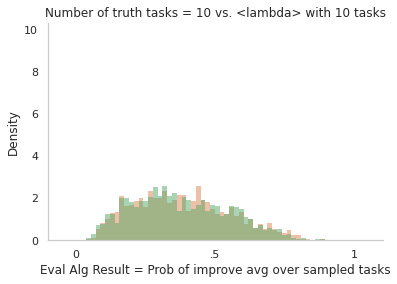}};
                            \node[above=of img3, node distance=0cm, xshift=0cm, yshift=-1.2cm,font=\color{black}] {\small{$10$ filtered tasks, $10$ holdout tasks}};
                        \end{tikzpicture}
                        \vspace{-0.015\textheight}
                        \caption{
                            We look at the distribution of evaluate change results -- i.e., the improvement probability given the change -- on both the randomly filtered tasks and the holdout tasks, as we vary the number of filtered and holdout tasks.
                            For all results we use $5$ quality samples.
                            \textbf{Takeaway:}
                            We should account for how we select the number of filtered tasks and holdout tasks when assessing filters.
                        }\label{fig:filter_eval_example}
                        \vspace{-0.025\textheight}
                    \end{figure}

                \subsubsection{Investigating how we evaluate filters}\label{sec:app_eval_filter_results}
                    \hphantom{a}
                    \begin{figure}[H]
                        \centering
                        \vspace{-0.015\textheight}
                        \begin{tikzpicture}
                            \centering
                            \node (img){\includegraphics[trim={.75cm .75cm .25cm .25cm},clip, width=.47\linewidth]{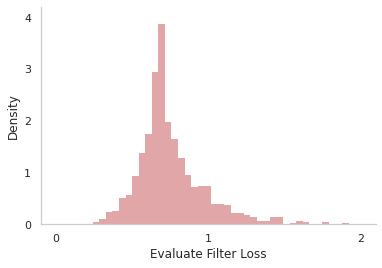}};
                            \node[left=of img, node distance=0cm, rotate=90, xshift=1.0cm, yshift=-.75cm, font=\color{black}] {Density};
                            \node[below=of img, node distance=0cm, xshift=3.35cm, yshift=1.15cm,font=\color{black}] {Filter Loss};
                            \node[above=of img, node distance=0cm, xshift=0cm, yshift=-1.4cm,font=\color{black}] {Change $=$ Default $\to$ DNN Only};
                            
                            \node [right=of img, xshift=-1cm](img2){\includegraphics[trim={1.25cm .75cm .25cm .25cm},clip, width=.45\linewidth]{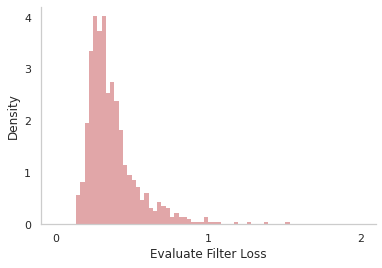}};
                            \node[above=of img2, node distance=0cm, xshift=0cm, yshift=-1.4cm,font=\color{black}] {Change $=$ Default $\to$ $5\times$ compute};
                        \end{tikzpicture}
                        \vspace{-0.015\textheight}
                        \caption{
                            We show the distribution of filter loss values from different samplings of train tasks and holdout tasks, for the changes in \Figure~\ref{fig:filter_eval_changes_and_mismatch}.
                        }\label{fig:filter_eval_loss_dist}
                        \vspace{-0.015\textheight}
                    \end{figure}

                \subsubsection{Comparing different filtering strategies}\label{sec:app_contrast_filter_results}
                    \hphantom{a}

                    \begin{figure}[H]
                        \centering
                        \vspace{-0.015\textheight}
                        \begin{tikzpicture}
                            \centering
                            \node (img){\includegraphics[trim={.75cm .75cm .0cm .75cm},clip, width=.48\linewidth]{images/filter_analyze_diffFilters_toDNN.png}};
                            \node[left=of img, node distance=0cm, rotate=90, xshift=2.75cm, yshift=-1cm, font=\color{black}] {\small{Loss Diff.~from Random Baseline}};
                            \node[below=of img, node distance=0cm, xshift=3.5cm, yshift=1.15cm,font=\color{black}] {Number of Tasks Selected by Filter};
                            \node[above=of img, node distance=0cm, xshift=0cm, yshift=-1.4cm,font=\color{black}] {\small{Change $=$ Default $\to$ DNN Only}};
                            
                            \node [right=of img, xshift=-1cm](img2){\includegraphics[trim={1.5cm .75cm .0cm .75cm},clip, width=.44\linewidth]{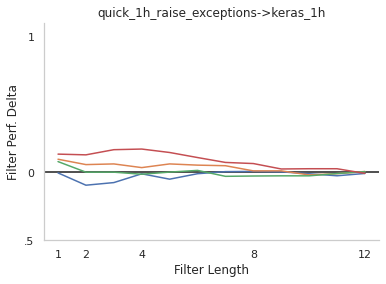}};
                            \node[above=of img2, node distance=0cm, xshift=0cm, yshift=-1.4cm,font=\color{black}] {\small{Change $=$ Default $\to$ Diff.~AutoML System}};
                        \end{tikzpicture}
                        \vspace{-0.015\textheight}
                        \caption{
                            The mean of the differences between a filter strategy with a random baseline -- multiple filter strategies are displayed in each color.
                        }\label{fig:analyze_filter_example_diffFilters}
                        \vspace{-0.015\textheight}
                    \end{figure}
                
                    \begin{figure}[H]
                        \centering
                        \begin{tikzpicture}
                            \centering
                            \node (img){\includegraphics[trim={.75cm .75cm .0cm .75cm},clip, width=.48\linewidth]{images/filter_analyze_diffChanges_true.png}};
                            \node[left=of img, node distance=0cm, rotate=90, xshift=2.75cm, yshift=-1cm, font=\color{black}] {\small{Loss Diff.~from Random Baseline}};
                            \node[below=of img, node distance=0cm, xshift=3.5cm, yshift=1.15cm,font=\color{black}] {Number of Tasks Selected by Filter};
                            \node[above=of img, node distance=0cm, xshift=0cm, yshift=-1.4cm,font=\color{black}] {\small{Approximate Oracle Similarity Filter}};
                            
                            \node [right=of img, xshift=-1cm](img2){\includegraphics[trim={1.5cm .75cm .0cm .75cm},clip, width=.44\linewidth]{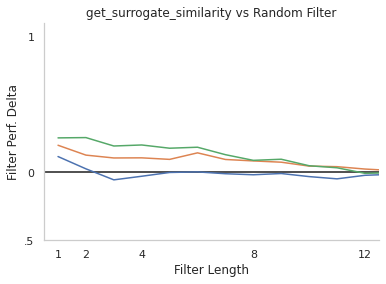}};
                            \node[above=of img2, node distance=0cm, xshift=0cm, yshift=-1.4cm,font=\color{black}] {\small{Performance Descriptor Similarity Filter}};
                        \end{tikzpicture}
                        \caption{
                            The mean of the differences between a filter strategy with a random baseline -- multiple system changes are displayed in each color.
                            \textbf{Takeaway:}
                            The approximate oracle similarity filter always improves performance over the random baseline.
                            The performance descriptor similarity improves performance over the random baseline when assessing changing to the search space, but not for changes to the implementation.
                        }\label{fig:analyze_filter_example_diffChanges}
                    \end{figure}
                
                    \begin{figure}[H]
                        \centering
                        \begin{tikzpicture}
                            \centering
                            \node (img){\includegraphics[trim={.75cm .75cm .0cm .75cm},clip, width=.48\linewidth]{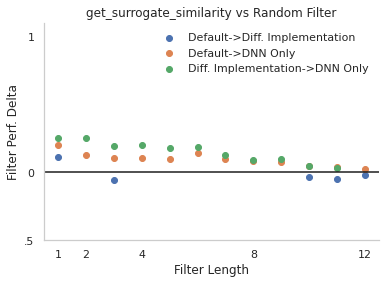}};
                            \node[left=of img, node distance=0cm, rotate=90, xshift=2.75cm, yshift=-1cm, font=\color{black}] {\small{Loss Diff.~from Random Baseline}};
                            \node[below=of img, node distance=0cm, xshift=3.5cm, yshift=1.15cm,font=\color{black}] {Number of Tasks Selected by Filter};
                            \node[above=of img, node distance=0cm, xshift=0cm, yshift=-1.4cm,font=\color{black}] {Performance Descriptor Filter};
                            
                            \node [right=of img, xshift=-1cm](img2){\includegraphics[trim={1.5cm .75cm .0cm .75cm},clip, width=.44\linewidth]{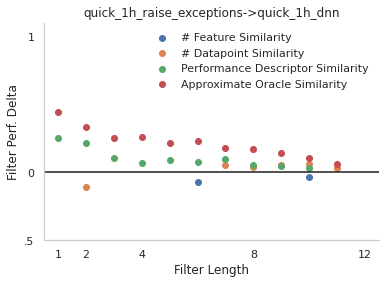}};
                            \node[above=of img2, node distance=0cm, xshift=0cm, yshift=-1.4cm,font=\color{black}] {Change $=$ Default $\to$ DNN Only};
                        \end{tikzpicture}
                        \caption{
                            The mean of the differences in quality, relative to a random baseline, and filtered to only show values with a significance level greater than $.05$ between a filter strategy with a random baseline -- for different changes in setups.
                            \textbf{Takeaway:}
                            The significance filtered results are similar to their respective unfiltered results in \Figure~\ref{fig:analyze_filter_example_diffFilters} and \ref{fig:analyze_filter_example_diffChanges}.
                        }\label{fig:analyze_filter_example_significance}
                    \end{figure}
                

\end{document}